\documentclass[a4paper,times]{article}
\pdfoutput=1
\usepackage[utf8]{inputenc}

\usepackage{graphicx}
\usepackage{subcaption}
\usepackage{booktabs}
\usepackage{multirow}
\usepackage{amsmath}
\usepackage{amssymb}
\usepackage{url}
\usepackage{lineno}

\begin{document}

\title{Robotic manipulation of multiple objects as a POMDP}
\date{}
\author{Joni Pajarinen, Joni.Pajarinen@aalto.fi \and
  Ville Kyrki, Ville.Kyrki@aalto.fi\\
  Department of Electrical Engineering and Automation,\\
  Aalto University, Finland}
\maketitle

\begin{abstract}
  This paper investigates manipulation of multiple unknown objects in
  a crowded environment. Because of incomplete knowledge due to
  unknown objects and occlusions in visual observations, object
  observations are imperfect and action success is uncertain, making
  planning challenging. We model the problem as a partially observable
  Markov decision process (POMDP), which allows a general reward based
  optimization objective and takes uncertainty in temporal evolution
  and partial observations into account. In addition to occlusion
  dependent observation and action success probabilities, our POMDP
  model also automatically adapts object specific action success
  probabilities. To cope with the changing system dynamics and
  performance constraints, we present a new online POMDP method based
  on particle filtering that produces compact policies. The approach
  is validated both in simulation and in physical experiments in a
  scenario of moving dirty dishes into a dishwasher. The results
  indicate that: 1) a greedy heuristic manipulation approach is not
  sufficient, multi-object manipulation requires multi-step POMDP
  planning, and 2) on-line planning is beneficial since it allows the
  adaptation of the system dynamics model based on actual experience.
\end{abstract}

\section{Introduction}
For a service robot, physical interaction with its environment is an
essential capability. As the application areas of service robotics are
extending to complex unstructured environments, robotic manipulation
has become an important focus area within robotics research.

In complex environments, the robot's knowledge about its environment
is incomplete and uncertain. To operate in such environments, robots
can employ different mechanisms. First, a robot may use on-line
sensing to attempt to gain more information about the
environment. Second, sensory measurements can be used directly in a
feedback control loop to adapt to small disturbances. Third, the
uncertainty can be taken into account on the level of planning the
actions.

Planning under uncertainty with imperfect sensing can be modelled as a
partially observable Markov decision process (POMDP). While POMDPs
have been one of the actively pursued research directions within AI,
they have not been applied very widely in robotic manipulation. This
is partly because the manipulation planning problems have intrinsic
characteristics such as continuous state spaces which make the
application of POMDP solvers less straightforward, and partly because
the manipulation planning approaches have only recently advanced to
the point where the explicit modeling of uncertainty becomes
tractable. The question which robotic manipulation problems benefit
from the explicit modeling of uncertainty in a POMDP framework remains
open to a large extent.

In this paper, we consider a multi-object manipulation planning
problem with the environment state estimated by imperfect sensors. The
dynamics of the system are considered to be partly unknown, which
supports the goal of long term autonomy of the robotic system. Our
high-level research question is in which situations explicit planning
under uncertainty is beneficial. The manipulation planning problem is
considered on task level, that is, the desired result of planning is
the best action to be performed. While motion planning for an
individual action, such as grasping an object, is out-of-the-scope for
this paper, the relationship of such an individual action to
observations of the overall system state and the world model dynamics
are modeled and learned during operation.

The theoretical contributions of the paper are two-fold: first, a
POMDP model for multi-object manipulation is proposed. As a particular
novel contribution, the model considers the effect of visual
occlusions on observations and success of actions. Moreover, the
dynamics of the world model, in particular related to the success of
actions, are updated during operation as more information is
gained. Second, we propose a new POMDP method which is applicable to
manipulation planning. In particular, it does not require a discrete
world model but instead samples the world model to construct
policies. The method is also scalable, does not require heuristics,
can handle uncertainty in the world model, and allows online planning,
which is important when the world model is not accurate. Furthermore,
the method produces compact policies in a predefined time. This can be
beneficial in a robotic setting where easy to inspect policies may
give new insights into the problem.

The proposed approaches are experimentally evaluated both in
simulation and with a real robot. The experiments demonstrate that
multi-step, longer horizon planning is beneficial in complex
environments with clutter. In particular, POMDPs are beneficial if a
particular problem has some of the following characteristics: 1) the
problem requires weighting the value of information gathering versus
collecting immediate rewards such as lifting objects to get a better
view on other objects, 2) the world model is uncertain and thus it
should be updated, for example when some objects are harder to grasp
than others, or 3) the sequence of actions matters such as when
objects occlude each other even partially. Altogether, the paper is
the first to propose long term POMDP planning for manipulating many
objects in a high dimensional, unknown, and cluttered environment.
   
\section{Related work}

\subsection{Partially observable Markov decision processes}

A partially observable Markov decision process
(POMDP)~\cite{kaelbling98} defines the optimal policy for a sequential
decision making problem while taking into account uncertain state
transitions and partially observable states. This makes POMDPs
applicable to diverse application domains such as
robotics~\cite{thrun02}, elder care~\cite{hoey07}, tiger
conservation~\cite{chades08}, and wireless
networking~\cite{pajarinen09}. However, versatility comes with a
price, the computational complexity of finite-horizon POMDPs is
PSPACE-complete~\cite{papadimitriou87} in the worst case.

Because of the high computational complexity, state-of-the-art POMDP
methods~\cite{smith05,kurniawati08,silver10,bai11,shani13} use
different kinds of approximations. There are at least two causes for
the intractability of POMDPs: 1) state space size, and 2) policy
size. State-of-the-art POMDP methods yield good policies even for
POMDP problems with hundreds of thousands of
states~\cite{smith05,kurniawati08} by trying to limit policy search to
state space parts that are reachable and relevant for finding good
policies. However, in complex real-world problems the state space can
be still much larger. In POMDPs with discrete variables, the state
space size grows exponentially w.r.t\ the number of state
variables. In order to make POMDPs with large state spaces tractable,
there are a few approaches: compressing probability distributions into
a lower dimension \cite{poupart02}, using factored probability
distributions \cite{mcallester99, pajarinen10}, or using particle
filtering to represent probability distributions
\cite{silver10,bai11}. Particle filtering is particularly attractive,
because an explicit probability model of the problem is not needed. In
fact, in order to cope with a complex state space, we use particle
filtering in the online POMDP method presented in more detail in
Section~\ref{sec:pomdp_method}.

Assuming the problem of state space size solved, the problem of policy
size still remains. In the worst case, the size of a POMDP policy
grows exponentially with the planning horizon. Offline POMDP methods
\cite{shani13} take advantage of the piecewise linear convexity (PWLC)
of the POMDP value function to keep the size of the policy reasonable.
Some offline POMDP methods use fixed size policies. A common approach
is to use a fixed size (stochastic) finite state controller
\cite{poupart03,amato10a} as a policy. The monotonic policy graph
improvement method \cite{pajarinen11c} utilizes a fixed size \emph{policy
graph}, an idea which is also adopted here.

Contrary to offline approaches, online POMDP methods \cite{ross08b}
compute a new policy at each time step. Online planning starts from
the current belief and can thus concentrate on only the part of the
search space that is currently reachable. Moreover, restarting
planning from the current belief allows the online planner to correct
planning ``mistakes'', which an inaccurate world model caused in
earlier time steps. This is especially relevant in robotic
manipulation in service settings where an accurate world model is
difficult to estimate for example due to unknown objects. Online POMDP
methods usually represent the policy as a policy tree. Techniques such
as pruning can be used in order to reduce the size of the policy tree,
but this does not solve the problem of exponential growth of the
policy tree w.r.t.\ the planning horizon.
The online POMCP method of Silver et al.\ \cite{silver10} uses
particle filtering to address the state space size problem and
Monte-carlo tree search to explore the policy space. However, for best
results, POMCP requires a problem specific heuristic
\cite{silver10}. POMCP is also not designed to produce compact
policies. The online POMDP approach that we propose allows for long
planning horizons by using a compact, fixed-size policy graph
\cite{pajarinen11c}. Because of the compact policy size, the policy
can be inspected by a domain expert.

In manipulation, and more generally in robotics, the world model is
often uncertain and thus it is necessary to learn the world model
during online operation. The goal then is to maximize total reward
while taking into account that the current world model is not accurate
and that actions can yield information about the world model. For a
discrete POMDP, a natural Bayesian approach is to model transition and
observation probabilities with Dirichlet distributions
\cite{doshi08,poupart08,ross11}. Following this, our probability model
uses Beta distributions to model uncertainty in object specific grasp
probabilities. Few papers \cite{ross08c,bai13} exist on using a POMDP
model with uncertain probabilities in robotics. Ross et
al. \cite{ross08c} apply an online POMDP approach to simulated robot
navigation with Gaussian distributions with unknown parameters. Bai et
al. \cite{bai13} present an offline POMDP planning approach for robot
motion planning with unknown model parameters and validate their
approach in simulation. However, we are not aware of prior robotic
manipulation research that uses a POMDP model with uncertain
probabilities.

\subsection{Manipulation under uncertainty}
Manipulation planning under uncertainty is not a new problem to be
considered in robotics. Already in the early 1980's, Lozano-P\'eres et
al.\ \cite{Lozano-Peres-ijrr1984} considered the automatic synthesis
of fine-motion actions under initial robot pose uncertainty using
compliant motion, preimages and backward chaining. The originally
sensorless decision-theoretic line of research can be seen to continue
to this day with extensions over the years to e.g. grasping
\cite{Brost-ijrr1988} and a probabilistic setting
\cite{LaValle-ijrr1998}. A good summary of the current state in this
line of work can be found in \cite{LaValle-2006}.

There is a recent trend to integrate task and motion planning, see
\cite{Kaelbling-mitcsail2012-018} for an overview. However, these
approaches do not address directly the problem of uncertainty. The
only exception is the recent work by Kaelbling and Lozano-Peres
\cite{Kaelbling-mitcsail2012-019}, where preimage backchaining is used
for belief-space planning in a hierarchical framework. The approach
handles probabilistic uncertainty by using a deterministic
approximation of the domain and replanning after each time step. Our
work differs from this approach: we consider the interactions of the
manipulation actions, that often occur in multi-object manipulation,
as well as update the world model based on on-line experience.

There is a tradition to formulate robot navigation problems as POMDPs,
for an overview see \cite{Thrun-2005}, or a recent study for long time
horizon POMDP planning \cite{Kurniawati-ijrr2011}. In manipulation,
the use of the formulation is not common. Hsiao et
al.\ \cite{Hsiao-icra2007} proposed the partitioning of the
configuration space of grasping with one uncertain degree of freedom
to yield a discrete POMDP which can be solved for an optimal
policy. In grasp planning, the state-of-the-art includes probabilistic
approaches with a short time horizon. The goal can be formulated
either as positioning the robot accurately as in \cite{Hsiao-ar2011}
or maximizing the probability of a successful grasp as in
\cite{Hsiao-icra2011ws,Laaksonen-iros2012}. The short-term planning
can also be extended to include information gathering actions
\cite{Nikandrova-ras2013}. In contrast to the above, this paper
considers manipulation of multiple objects which are unknown and where
the sequence of actions has a significant effect.

Recent work by Dogar and Srinivasa \cite{Dogar-ar2012} proposes
manipulation of multiple objects using grasping and pushing
primitives. The approach uses pushing to collapse the uncertainties of
the object locations as well as to clear clutter in the scene. The
planning is performed at the level of object poses. Monso et al.\
\cite{Monso-iros2012} proposed to formulate clothes separation as a
POMDP. In contrast to our work, the approach of Monso et al.\ is
environment specific. Monso et al.\ rely on a clothes separation
specific state space definition, which models the number of clothes in
each area. We model object attributes, associated probabilities, and
grasp probabilities, in any kind of environment.

\section{Multi-object manipulation: a POMDP}

In multi-object manipulation, a robot performs actions on several
objects. In particular, the robot may grasp objects, move them, or use
them in another way to accomplish some predefined goals. In this
paper, we focus on the problem of deciding how to manipulate unknown
objects in a crowded environment. Because the environment is crowded,
only parts of the objects can be observed by visual sensors. In
addition to uncertain observations, real-world manipulation problems
have uncertain action consequences, especially when the robot does not
have a model of the objects beforehand, or when the robot does not
observe the objects well. For example grasping or moving an object may
fail, because the shape or location of the object differs from the
observed one. Real-world problems often have several (possibly
conflicting) goals. As a practical example consider putting dirty
dishes from a table full of dishes into a dishwasher: the goal is to
maximize the number of dirty dishes in the dishwasher, minimize the
number of clean dishes in the dishwasher, and minimize the execution
time. In order to address the issue of complex objectives, uncertain
observations, and uncertain action effects, this paper models the
problem of manipulating multiple unknown objects as a partially
observable Markov decision process (POMDP). 

Planning of manipulation, for instance grasp planning, is
traditionally considered as a geometrical problem. However, in
unstructured environments with unknown objects the current
state-of-the-art approaches often plan individual actions (e.g., a
grasp) directly based on the observed environment
\cite{Fischinger-icra2013}.  We follow the same idea so that the
planning is performed on the level of semantic actions and locations,
while the execution of individual actions is then performed based on
the currently observed scene. However, our approach also models the
interplay of the immediate sensor measurements to both observation and
system models. For example, the rate of visual occlusion modulates
the probability of correct observations and successful completion of
actions.

We begin by defining a POMDP, then describe a new online POMDP
planning method which is suitable for complex problems such as
multi-object manipulation, and finally describe how to model
multi-object manipulation in crowded environments as a POMDP with an
application of moving dirty dishes into a dishwasher.
  
\subsection{What is a POMDP?}

A POMDP is a model that defines optimal behavior for a given Markovian
problem, taking into account uncertainty in observations as well as
action effects over a potentially long time horizon. In a POMDP, a set
of hidden Markov models, one for each action choice, describes the
temporal dynamics of the problem and the optimization objective is
defined by assigning a reward to each action in each possible
situation. In a specific application, rewards should reflect real
value, e.g.\ monetary cost.

Formally a POMDP is defined by the tuple $\langle \mathcal{S},
\mathcal{A}, \mathcal{O}, P, R, O, b_0 \rangle$, where $\mathcal{S}$
is the set of states, $\mathcal{A}$ is the set of actions, and
$\mathcal{O}$ is the set of observations. The state set includes all
possible states of the world, in which the agent is assumed to operate
in. $P(s^{\prime} | s, a)$ is the transition probability to move from
state $s$ to the next time step state $s^{\prime}$, when action $a$ is
executed. $R(s, a)$ yields the real-valued reward for executing action
$a$ in state $s$ and $O$ denotes the observation probabilities $P(o |
s^{\prime}, a)$, where $o$ is the observation made by the agent, when
action $a$ was executed and the world moved to the state
$s^{\prime}$. Lastly, $b_0(s)$ is the initial state probability
distribution, also known as the initial \emph{belief}. In a
finite-horizon POMDP, the goal is to optimize the expected reward
\begin{equation}
  E\left[ \sum_{t=0}^{T-1} R(s(t), a(t)) | \pi \right] \;,
  \label{eq:finite_horizon_POMDP}
\end{equation}
where $T$ is the horizon, $s(t)$ is the state, and $a(t)$ the action
chosen at time step $t$ by the policy $\pi$.

Because the states are not fully observable, the current state
cannot be used for decision making as in fully observed
models. Instead, the \emph{belief} $b(s)$, a
probability distribution over world states, is maintained to make (optimal)
decisions at each time step. Starting from the initial belief $b_0(s)$, 
the belief is updated at each time step. After performing action $a$
and observing $o$ the updated belief $b^{\prime} =
b^{\prime}(s^{\prime} | b, a, o)$ can be obtained from the current
belief $b = b(s)$ using the Bayes formula $b^{\prime}(s^{\prime} | b,
a, o) = \frac{P(o | s^{\prime}, a)}{C} \sum_s P(s^{\prime} | s, a)
b(s)$, where $C$ is a normalizing constant.

To give an intuitive idea of how POMDPs can be applied in practice, we
will now give short examples for the transition and observation
probabilities, and the reward function. In a POMDP, the transition
probability $P(s^{\prime} | s, a)$ models the uncertainty in action
effects: what is the probability to move a cup successfully from a
table (part of state $s$) into a dishwasher (part of $s^{\prime}$),
when the action $a$ is ``move cup into dishwasher''? The observation
probability $P(o | s^{\prime}, a)$ models the uncertainty in
observations: what is the probability of observing a cup as dirty
(observation $o$), when it is dirty (part of state $s^{\prime}$) and
we are executing action $a$ ``look at cup''? Finally, the reward
$R(s,a)$ explicitly specifies the optimization goal: gain positive reward for
moving (action $a$) a dirty cup (part of state $s$) into the dishwasher.

\subsection{Online policy graph POMDP using monotonic value improvement}
\label{sec:pomdp_method}

In this paper, as often in robotic applications, the state space of
the robotic manipulation task is high dimensional. The state space has
exponential size in the number of discrete state variables, and
includes uncertain grasp success probability distributions. Because of
the complex state space, POMDP methods based on exact probability
representations are not applicable. We present a new online POMDP
method based on the monotonic policy value improvement algorithm
\cite{pajarinen11c} proposed by us earlier. The next subsection
briefly introduces the method from \cite{pajarinen11c} followed by the
extensions: 1) the new method uses particle filtering to represent
probability distributions and estimate values in a way that takes
advantage of the policy graph (Sec.~\ref{sec:particle_filtering}),
instead of using a discrete tabular probability distribution
representation~\cite{pajarinen11c}, and 2) the new method is
transformed from an offline~\cite{pajarinen11c} method into an online
method in a policy graph specific way (Sec.~\ref{sec:offline_to_online}).

\subsubsection{POMDP policy and method}

\begin{figure}[htb]
  \centering
  \includegraphics[width=10cm]{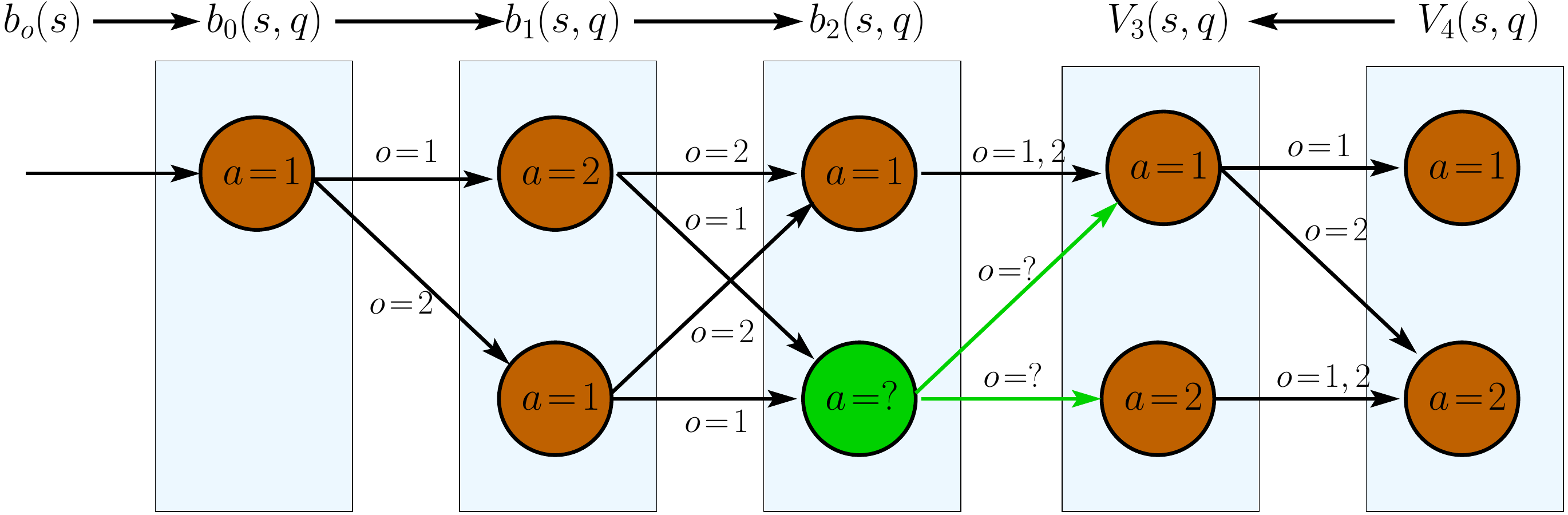}
  \caption{Illustration of a policy graph node update in the monotonic
    policy graph value improvement algorithm for POMDPs.}
  \label{fig:policy_graph_algorithm}
\end{figure}

We represent the policy of the agent (the robot) as a policy graph $G$
(see Fig. \ref{fig:policy_graph_algorithm} for an example policy
graph) beginning from time step $t = 0$ and ending at the planning
horizon $t = T - 1$. Each graph node defines a conditional plan for
the robot to follow: which action to perform, and depending on the
observation made, to which next layer node to transition next. 

The policy improvement approach in \cite{pajarinen11c} uses dynamic
programming to improve each policy graph layer at a time. First, the
approach computes the belief $b_t(s,q)$ at each layer $t$ and each
graph node $q$ starting from the initial belief $b_0(s)$ in the first
layer. Then, starting from the last layer and moving one layer at a
time towards the first layer, the approach computes for each node in
the layer a new policy (action, observation edges), which maximizes
the expected reward for the belief at the current node. The expected
reward is computed from the immediate reward and the next layer value
function $V_{t+1}(s, q)$, which yields the expected reward when
starting from state $s$ and graph node $q$ in layer $t+1$ and
following the policy graph until layer $T-1$. This procedure
guarantees monotonic improvement of policy value. For algorithmic
details, see \cite{pajarinen11c}.

In order to keep computations tractable, we use a policy graph with
fixed width and depth. This circumvents the problem of exponential
growth of a search tree, allows for manual inspection of a compact
policy, and enables us to convert the offline approach to an online
one.

\subsubsection{Particle filtering}
\label{sec:particle_filtering}

The method in \cite{pajarinen11c} assumes a discrete ``flat''
POMDP. In order to deal with a large state space, we use particle
filtering to approximate beliefs and for estimating values.

\textbf{Belief representation and update.} We represent a belief
$b(s)$, a probability distribution over $s$, as a finite set of
particles~\cite{Thrun-2005}, that is, a weighted set of state
instances $s^j$. The belief is $b(s) = \sum_j w^j \delta (s, s^j);
\sum_j w^j = 1; 0 \le w^j \le 1$, where $w^j$ is the particle weight
and $\delta (s, s^j) = 1$ when $s = s^j$ and zero otherwise. What a
state actually is depends on the application: Section
\ref{sec:manipulation_as_POMDP} defines a state for multi-object
manipulation.

We use two kinds of belief updates. The first one is the commonly used
update of the current belief $b(s)$, when an action has been executed
and an observation made. This belief update is used in initializing
the policy graph and for sampling new beliefs for redundant policy
graph nodes in order to re-optimize them (if the optimized policy at a
policy graph node, that is, the action and connections to the next
layer, is identical to the policy of another node in the layer, we
sample a new belief over world states, and re-optimize the node for
the new belief. This ``compresses'' the policy graph without changing
its value \cite{pajarinen11c}). The second kind of belief update
projects the belief $b_t(s, q)$ to the next layer belief $b_{t+1}(s,
q)$, using the current policy. The second belief update is used in
each improvement round.

In the first belief update, the action $a(t)$ and observation $o(t+1)$
are given. We sample a next time step state $s^j(t+1)$ for each
current state $s^j(t)$ according to the application specific dynamics
(state transition) model $P(s^j(t+1) | s^j(t), a(t))$. We then compute
the new particle weight $w^j(t+1)$ as the product of the old weight
and the observation probability: $w^j(t+1) = w^j(t) P(o(t+1) |
{s^j}(t+1), a(t))$. As usual, to prevent particle impoverishment, we
resample particles, when the effective sample size drops below a
threshold ($0.1$ in the experiments).

In the second belief update, for updating the belief $b_t(s, q)$, a
particle consists of a weight $w^j(t)$ and a state/node pair $(s^j(t),
q^j(t))$. To sample a new particle $(s^j(t+1),q^j(t+1))$ using a
$(s^j(t), q^j(t))$ pair, we first get the action $a(t)$ for node
$q^j(t)$. Then, we sample a new state $s^j(t+1)$ from $P(s^j(t+1) |
s^j(t), a(t))$. Next, we sample an observation $o(t+1)$ from $P(o(t+1)
| s^j(t+1), a(t))$. Finally, the observation edge for observation
$o(t+1)$ of the graph node $q^j(t)$ yields the new graph node
$q^j(t+1)$. In this update, the particle weights do not change.

As a side remark, note that our approach differs from existing
particle filtering based approaches. In order to improve the policy,
we use the current policy for finding a belief distribution over graph
nodes, but other state-of-the-art POMDP methods based on particle
filtering \cite{silver10,bai11} select an action and observation to
find a new belief for which to compute a policy. In other words, other
POMDP methods use a constant amount of particles to represent a single
belief, but we use a constant amount of particles to represent the
belief over a policy graph layer (a time step) and each graph node is
assigned particles proportional to the probability of the graph
node. We believe this will result in a more efficient use of the
computational resources.

\textbf{Value estimation.} In order to determine the best action and
observation edges for a policy graph node, the method has to estimate
the value for each action-observation-next node triplet. From these
triplets the method can then select for each action the highest value
observation-next node pairs and based on these select the highest
value action. To do this efficiently, we follow Algorithm 1 in
\cite{bai11}. The algorithm samples state transitions and observations
for each action and for the sampled observation simulates the value
for each next controller node. Bai et al.\ \cite{bai11} represent the
policy as a possibly cyclic finite state controller, but we use
instead an acyclic policy graph. However, no significant modifications
are necessary because the algorithm is based on
simulation. Furthermore, the bound for the approximation error induced
by sampling, shown in Theorem 1 in \cite{bai11}, also applies here:
the error is bounded by a term that decreases at the rate of
$\mathcal{O}(1/\sqrt{N})$, where $N$ is the number of samples.

In the implementation we do not actually sample states from a belief,
but just go through all particles, one at a time, and utilize the
particle weight for value estimation. In the policy improvement round,
this is more efficient than sampling states, because particles usually
have identical weights.

\textbf{Complexity.} The worst case complexity of one policy
improvement round of the POMDP method is quadratic w.r.t.\ the
planning horizon because the method simulates state trajectories up to
the planning horizon for each policy graph layer. In the experiments
in Section~\ref{sec:experiments}, the method performed well. In the
future, one could parallelize the algorithm to utilize multiple CPU
cores (easily because of the particle representation of
probabilities), or use a fixed sampling depth.

\subsubsection{From offline to online}
\label{sec:offline_to_online}

Because of the computational and modeling restrictions discussed
previously, we transform the offline POMDP method into an online
one. Similarly to the receding horizon control (RHC) approach
\cite{mattingley11,chakravorty11} in automatic control we re-plan at
each time step up to a finite horizon. Intuitively, we use a moving
window that at each time step shifts one step to the right over the
policy graph (imagine this with the help of the policy graph in
Fig. \ref{fig:policy_graph_algorithm}), discards the first layer, and
adds a new layer at the end. At the beginning, the agent optimizes the
policy graph for several improvement rounds for the initial
belief. Then in following time steps the agent estimates the new
belief and constructs a new policy for the belief, as follows: 1)
initialize the new policy graph with the layers $2, \dots, T - 1$ of
the previous policy graph; 2) add a new last layer to the policy graph
with random actions, and add random observation edges to the layer
preceeding the last layer; 3) use the regular policy graph improvement
method on the new policy graph.  The basic idea here is to initialize
the current policy graph using the policy graph of the previous time
step, and then optimize the policy graph for the current
belief. Because of the initialization, the required number of
improvement rounds during online operation is then less when compared
to offline optimization.

\subsection{Multi-object manipulation as a POMDP}
\label{sec:manipulation_as_POMDP}

We discuss now a general POMDP framework for modeling multi-object
manipulation. Later, in Section~\ref{sec:dirty_dishes}, we then show
how the POMDP framework can be applied to the problem of moving dirty
dishes into a dishwasher.

In multi-object manipulation, the robot has to decide at each time
instance which object to manipulate. We consider problems, where the
world consists of $N$ objects with varying attributes. The total
number of actions is $\sum_i |A_i|$, where $|A_i|$ denotes the number
of possible actions for object $i$. In each time step, the action of
the robot changes the spatial locations and poses of the objects, and
the robot makes an observation about the changed state of the
world. Our POMDP model uses discrete actions and
observations. However, instead of forcing the robotic planning problem
into a manageable discrete state space as is done e.g.\ in
\cite{Monso-iros2012}, we use a POMDP method based on particle
filtering (discussed in Sec.~\ref{sec:pomdp_method}) that allows us to
maintain complex object information required for efficient
multi-object manipulation.

\textbf{State space and actions.} The state space consists of semantic
object locations (e.g. ``on table'', ``in a dishwasher''), object
attributes, and historical data of observations and action successes
for each object. The model assumes that the semantic location of an
object is constant over time unless a manipulation action successfully
changes it. However, because an online planning approach is used, the
planning always restarts from the current belief taking into account
the most recent measurements.

Formally, the POMDP state $s = (s_1, s_2, \dots, s_N)$ is a
combination of object states $s_i = (s_i^{\textrm{loc}},
s_i^{\textrm{attr}}, s_i^{\textrm{hist}})$ where $s_i^{\textrm{loc}}$
is the semantic object location, $s_i^{\textrm{attr}}$ the object
attributes, and $s_i^{\textrm{hist}}$ compressed historical
information of action successes and object attribute observations. The
action success information consists of a count of succeeded
$n_i^{\textrm{succ}}$ and failed $n_i^{\textrm{fail}}$ grasps for each
object. Because of the finite number of objects the number of action
counts is finite. Similarly, as discussed in more detail below, the
number of different object attribute observations is
finite. Therefore, $s_i^{\textrm{hist}}$ has finite dimensionality,
and the POMDP state can be stored and operated on efficiently. Note
that the POMDP states have the Markov property because the probability
for the next state depends only on the current state (and action).

The observation history contains information of past observations of
object attributes. Past object attribute observations can be used to
compute the probability distribution over an object's
attributes. Additionally, these are needed during planning because
future observations of the attributes can not be assumed statistically
independent, because the main source of observation uncertainty is
occlusion. In contrast, unless the occlusion changes, we assume that
an identical observation of the attribute is made (note that we assume
differently occluded observations independent). We assume that the
probability of making the correct observation depends on how occluded
the object is (we discuss this in more detail shortly). In more
detail, $s_i^{\textrm{hist}}$ contains the observation made in each
occlusion setting. For example, in the experiments objects can be
temporarily lifted: in addition to the current occlusion setting, we
store the observation for each object which was temporarily lifted and
which is otherwise in front of the observed object. Note that because
of the finite number of objects the number of occlusion settings is
finite, and thus the observation history has finite size.

\begin{figure}[htb]
  \centering
  \begin{tabular}{ll}
    \includegraphics[width=5cm]{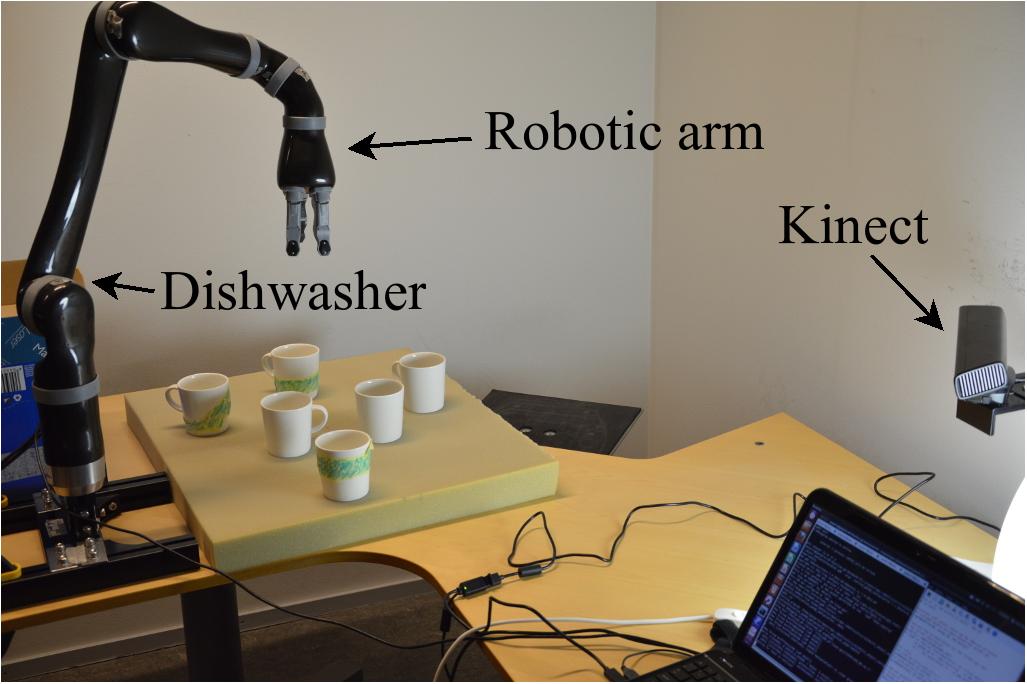} &
    \includegraphics[width=5cm]{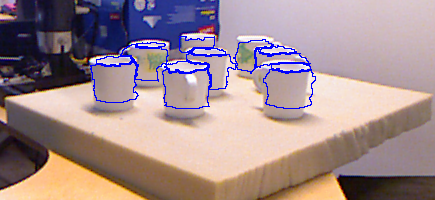}
  \end{tabular}
  \caption{Experimental setup. \textbf{Left:} A Kinova Jaco robotic
    arm manipulates objects placed on the table. A Microsoft XBOX
    Kinect acts as a monocular visual sensor for capturing RGB-D point
    clouds. In the experiments, the goal is to pick up dirty objects,
    here cups marked with a green color, from the table and place them
    into the ``dishwasher'', represented by the blue box on the far
    left. \textbf{Right:} An image captured by the Kinect
    sensor. Object edges are depicted in blue.}
  \label{fig:overall_experimental_setup}
\end{figure}

\textbf{Occlusion ratio.} The action success probability and the
observation probability of an object depend on how occluded the object
is. Because we do not have models for the objects, the occlusion is
modeled using a model free \emph{occlusion ratio}. The reasoning is
that the higher the occlusion, the smaller the probability of success
in actions or observations. In the experiments, we capture a point
cloud, segment the point cloud into objects, compute edges for all
objects using 2-D information, and then find out how much the edges of
objects touch each other. The right hand side figure in
Fig.~\ref{fig:overall_experimental_setup} shows edges found for
segmented objects in a scene. When the edge of object A, which is
closer to the visual sensor, touches the edge of object B, object A
occludes object B. 

Consider computing the occlusion ratio for object B. Denote with
$\textrm{TOT}$ the perimeter of the 2D contour of object B, that is,
the total number of 2D pixels for which the number of neighboring 2D
pixels, which are part of object B, is less than eight. Denote with
$\textrm{TOU}$ the touching edge between A and B, that is, the number
of 2D pixels in B which have atleast one neighboring 2D pixel in
object A (when B is occluded by several objects, just use the 2D
pixels of the occluding objects). The occlusion ratio for object B is
$1$, when $\textrm{TOT}$ subtracted by $\textrm{TOU}$ is smaller than
$\textrm{TOU}$, $0$ when $\textrm{TOU}=0$, and otherwise $\textrm{TOU}
/ (\textrm{TOT} - \textrm{TOU})$. The reasoning is that when an object
almost completely occludes another object, $\textrm{TOT}$ is roughly
double $\textrm{TOU}$. Thus an occlusion ratio of $1$ corresponds to
totally occluded and an occlusion ratio of $0$ to no occlusion at
all. The convenience variable $s_i^{\textrm{occl}}$ denotes the
occlusion ratio of object $i$.

In this paper, POMDP state transitions are based on sampling. When an
object A is sampled to be moved, so that it does not occlude another
object B anymore, it is straightforward to update the occlusion ratio
of B by removing the touching edge between A and B. However, if there
is an object C, which occludes A (edges of A and C touch), but not B,
and A is moved away, then there is a possibility that C could occlude
B after the removal of A. We call this \emph{occlusion
  inheritance}. For simplicity, we do not take occlusion inheritance
into account in the experiments and leave it as future work.

\textbf{Grasp probability.} We assume that occlusion affects the grasp
probability of all objects in a similar way, but, in addition, we
assume that each object has unknown properties that affect the grasp
probability of that specific object: we do not know beforehand what
kind of grasp properties each object has. For example a cup that has
fallen down may be harder to grasp, than another cup, which is
standing upright (see Fig.~\ref{fig:cup_falls_down} for an
example). The probability of a successful grasp is modeled as
\begin{multline}
  P(\textrm{grasp succeeded} | s_i^{\textrm{occl}}, s_i^{\textrm{hist}}) =
  E[p_i^{\textrm{succ}}] \\
  p_i^{\textrm{succ}} \sim Beta(p_i^{\textrm{succ prior}} n^{\textrm{prior}} + 
    n_i^{\textrm{succ}}, (1 - p_i^{\textrm{succ prior}}) n^{\textrm{prior}} +
    n_i^{\textrm{fail}}) \;,
  \label{eq:grasp_success_prob}
\end{multline}
where $p_i^{\textrm{succ prior}}$ is the occlusion ratio specific
grasp success prior probability and $n^{\textrm{prior}}$ is the
strength of the prior. In the experiments, we mapped the occlusion
ratio to the grasp success prior probability $p_i^{\textrm{succ
    prior}}$ using a simple exponential function
\begin{equation}
  p_i^{\textrm{succ prior}} = \exp(- \theta_{\textrm{G1}} s_i^{\textrm{occl}} + \theta_{\textrm{G2}}) \;,
  \label{eq:grasp_success_prior}
\end{equation}
where $\theta_{\textrm{G1}}$ and $\theta_{\textrm{G2}}$ are parameters
that can be experimentally estimated from object grasps, for example,
using two different occlusion ratios.

Note that we model the grasp probability as the mean of the Beta
distributed random variable $p_i^{\textrm{succ}}$. It would be
possible to use a more complex model during planning, in which one
would sample the grasp probability from the Beta distribution, but we
expect this would increase the number of particles needed for planning.

\textbf{Observations.} We assume that the semantic locations and
dependencies (which cup is in front of which cup) are fully observed
and that grasp success is also fully observed. At each time step the
agent observes whether the grasp succeeded and makes an observation
about object attributes. Using these observations, we can compute a
probability distribution over object attributes, which is needed for
sampling the initial POMDP belief and for displaying attribute
probabilities. Note that if grasp success or semantic locations are
not fully observed, then we can not estimate the initial POMDP belief
directly using grasp success and object attribute
observations. Instead, we could update at each time step an
(approximate) belief according to the current action and observation
and use that as the initial POMDP belief. However, in many
applications, including the dishwasher application further down,
semantic locations such as ``object on table'', ``object in
dishwasher'', and thus also grasp success, are fully observed.

As discussed earlier, we assume that the robot observes an object
identically unless the occlusion changes.  Denote with $o_i^j$, the
observation for object $i$ when in the $j$th occlusion setting, and
with $a_i^j$ the action performed when observing $o_i^j$, then the
attribute probability given the history is
\begin{align}
 &P(s_i^{\textrm{attr}} | o_i^1, \dots, o_i^M, a_i^1, \dots, a_i^M) = \nonumber \\
 &\;\;\; \frac{P(o_i^1, \dots, o_i^M | s_i^{\textrm{attr}}, a_i^1, \dots, a_i^M) 
  P(s_i^{\textrm{attr}} | a_i^1, \dots, a_i^M)}{ 
  P(o_i^1, \dots, o_i^M | a_i^1, \dots, a_i^M)} = \nonumber \\
 &\;\;\; P(s_i^{\textrm{attr}} | a_i^1, \dots, a_i^M) \prod_{j=1}^M P(o_i^j | s_i^{\textrm{attr}}, a_i^j) / 
  \sum_{s_i^{\textrm{attr}}} \prod_{j=1}^M P(o_i^j | s_i^{\textrm{attr}}, a_i^j) \;,
  \label{eq:attribute_probability}
\end{align}
where we assumed that observations are conditionally independent given
the object attributes, but if needed and computationally possible one
can use joint probabilities. We assume that attributes (e.g.\ color)
do not change over time, and thus actions do not influence object
attributes: $P(s_i^{\textrm{attr}} | a_i^1, \dots, a_i^M) =
P(s_i^{\textrm{attr}})$. In the experiments, we assumed
$P(s_i^{\textrm{attr}})$ is uniform.

\subsubsection{Dirty cups into dishwasher}
\label{sec:dirty_dishes}

We now demonstrate how the framework can be used to model the problem
of moving dirty cups from a table into a dishwasher as a POMDP
(another realistic application could be moving dishwasher-safe cups,
instead of dirty cups, into the dishwasher). In this problem, the
robot can gain more information of attributes by removing occlusions
and gain information about the object specific grasp probability
through successful and failed grasps.

\textbf{State space and actions.} In addition to the grasping and
observation history discussed in
Section~\ref{sec:manipulation_as_POMDP}, the world state consists of
the semantic location $s_i^{\textrm{loc}} = \{\textsc{table,
  dishwasher}\}$, and the attributes $s_i^{\textrm{attr}}$ of an object
include dirtyness $s_i^{\textrm{dirty}} = \{\textsc{clean, dirty}\}$. The
robot can perform three kinds of actions. The \emph{FINISH} action
terminates the robot actions and assigns a negative reward to dirty
dishes remaining on the table. The \emph{LIFT} action tries to lift an
object to expose the objects behind it and allows the agent to gather
more information about the occluded objects. A small negative reward
representing time cost is associated with the action. Note that the
action takes less time than moving the object into the dishwasher. The
\emph{WASH} action tries to move an object into the dishwasher (in the
experiments, a box). If the move succeeds, the state of the object
changes from \textsc{table} to \textsc{dishwasher}. If the move succeeds
and the moved object is dirty, then a large reward is obtained. If the
move succeeds and the object is clean, a large negative reward is
obtained. Failed grasps cause a small negative reward accounting for
the time cost. 
Note that when implementing the model, we can compute the reward for
the \emph{WASH} action as the expected next time step reward using the
grasp success probability, instead of deferring reward computation
until the grasp has happened in the next time step.

\textbf{Observations.} At each time step the agent observes whether
the grasp succeeded and the dirtyness of the $k$ nearest objects (in
the experiments $k = 2$) which were occluded by the moved cup. In
total, $2^{k + 1}$ possible observations. In the experiments, we model
the conditional probability of observing cup $i$ as dirty when it is
dirty with
\begin{equation}
  P(o_i = \textsc{dirty} | s_i^{\textrm{dirty}} = \textsc{dirty},
  s_i^{\textrm{occl}}) = \exp(- \theta_{\textrm{D1}}
  s_i^{\textrm{occl}} + \theta_{\textrm{D2}}) \;,
  \label{eq:obs_dirty_prob}
\end{equation}
where $\theta_{\textrm{D1}}$ and $\theta_{\textrm{D2}}$ are parameters
that can be, similarly to the grasp probability, experimentally
estimated from captured point clouds and object labels. The
probability of observing a cup as dirty when it is clean is modeled
identically with 
\begin{equation}
  P(o_i = \textsc{dirty} | s_i^{\textrm{dirty}} = \textsc{clean},
  s_i^{\textrm{occl}}) = \exp(- \theta_{\textrm{C1}} s_i^{\textrm{occl}}
  + \theta_{\textrm{C2}}) \;,
  \label{eq:obs_clean_prob}
\end{equation}
where parameters $\theta_{\textrm{C1}}$ and $\theta_{\textrm{C2}}$ are
also estimated in the same way.

\section{Experiments}
\label{sec:experiments}
The experiments follow the scenario described above. The scene is
observed by an RGB-D sensor (Microsoft Kinect) and a 6-DOF Kinova Jaco
arm with an integrated 3-fingered hand is used to manipulate the
objects. The objects belong to two classes: clean white cups and cups
with green ``dirt'' representing dirty
objects. Fig.~\ref{fig:overall_experimental_setup} illustrates the
experimental setup: to the left a picture of the setup, and to the
right an image captured by the Kinect sensor.

\textbf{Rewards.} The robot receives a reward at each time step. The
reward depends on the action executed and the current state of the
world. As discussed in Section \ref{sec:dirty_dishes}, the robot can
execute three different kinds of actions. 
The \emph{FINISH} action
terminates the problem and accumulates a reward of $-5$ for each dirty
cup on the table. Similarly, to limit experiment run times, after ten
time steps, the problem is terminated and a reward of $-5$ for each
dirty on the table given. The \emph{LIFT} action lifts an object up
and yields a reward of $-0.5$ for both failed and succeeded
grasps. The \emph{WASH} action moves an object into the dishwasher. If
the move succeeds, then the reward is $+5$ for a dirty object and
$-10$ for a clean object. For a failed move the reward is $-0.5$.
In our dishwasher application, there was no well determined
objective. Rewards were designed based on the researchers'
understanding of the application.

\textbf{Methods.} The \emph{POMDP planning} method described in
Section~\ref{sec:pomdp_method} is initialized by 10 offline policy
improvement rounds. Then, at each time step 4 improvement rounds for
the current belief are executed. To evaluate the benefit of
planning under uncertainty, the POMDP approach is also compared
against heuristic decision making: The \emph{heuristic manipulation}
method assumes that observations are accurate and deterministic. It
tries to move the dirty cup that has the highest grasp success
probability into the dishwasher. If no cup is observed dirty it
performs the FINISH action. In the experiments, we used two versions
of the heuristic method: one which updates grasp probabilities according
to the grasp success history and another which does not remember any grasp
history.

\textbf{Point cloud into a world model.} In the experiments, the
visual sensor captures a point cloud, from which we extract objects,
their color, and information on how they occlude each other. From
these we estimate grasp and observation probabilities and use these
probabilities to plan which action to perform. In more detail, first
the Kinect sensor captures an RGB-D point cloud of the visual
scene. Without using prior information we segment~\footnote{For
  segmentation we use organized multiplane segmentation and organized
  euclidean cluster extraction, part of the point cloud library
  \url{http://www.point clouds.org/}.} the point cloud into
objects. From the 2D-image, we determine the edge of each object and
how much it touches other objects' edges (see edges in the right hand
side image of Fig.~\ref{fig:overall_experimental_setup}). Using the
object edges, we compute, as discussed in Section
\ref{sec:manipulation_as_POMDP}, occlusion ratios. However, because of
occlusion, segmentation may produce multiple objects for one complete
object. Therefore, we merge objects that occlude each other and are
close (occlusion ratio above $0.5$ and centroid distance below $8$cm)
into one object and re-compute its occlusion ratio. Next, we compute
object specific grasp (Equation \ref{eq:grasp_success_prob}) and
observation probabilities (Equations \ref{eq:obs_dirty_prob} and
\ref{eq:obs_clean_prob}) using the occlusion ratios, observation
history, and initially estimated parameters. We set the grasp prior
count $n^{\textrm{prior}} = 0.5$. Finally, we make an observation if an
object is dirty or clean based on the distance of the object color to
precomputed color prototypes.

\textbf{Grasping.} Grasping an unknown object is performed by
executing a top grasp, closing fingers around the centroid of the
point cloud of the object to grasp, similar to
e.g.~\cite{Felip-ras2013}.

\textbf{Estimating initial parameters}. Before actual experimental
runs, we estimated experimentally the parameters of grasping and
observation probability functions defined in Equations
\ref{eq:grasp_success_prior}, \ref{eq:obs_dirty_prob}, and
\ref{eq:obs_clean_prob}. To estimate grasp parameters we attempted to
lift cups positioned on the table using the robot arm, both when the
cups were occluded and when not, and estimated grasp parameters
($\theta_{\textrm{G1}} = -0.904$, $\theta_{\textrm{G2}} = -0.087$)
from the recorded success rates. For the occluded case we used the
average occlusion ratio. We estimated observation function parameters
for dirty ($\theta_{\textrm{D1}} = -0.895$, $\theta_{\textrm{D2}} =
-0.087$) and clean ($\theta_{\textrm{C1}} = -0.193$,
$\theta_{\textrm{C2}} = 0.0$) cups similarly, but instead of the
lifting success rate, we used the observation success rate.

\subsection{Experiments with simulated dynamics}
\label{sec:simulated_dynamics}

In multi-object manipulation, robot actions may have far reaching
consequences: lifting first cup A and then cup B, may increase the
probability of cup C being observed dirty from low to high by exposing
it more fully. The robot has to consider at each time step, whether
the information gain from lifting a cup yields more reward in the long
run than executing an action which may yield higher immediate
reward. Of course, because of the uncertainty in actions and
observations, the real decision making problem can be even more
complicated than this simple example implies. Consequently, our
hypothesis is that a heuristic greedy manipulation approach is not
sufficient and that planning several time steps into the future is
needed. In order to study this hypothesis, we experimentally compared
heuristic manipulation and the proposed POMDP approach with different
planning horizons. Note that even though we simulate world
dynamics, we estimate the grasp and observation probabilities using
the physical robot arm and real observed occlusions. Moreover, we
estimate the occlusions and locations of objects from point clouds
captured by the Kinect sensor.

\begin{figure}[htb]
  \setlength{\tabcolsep}{5pt}
  \begin{tabular}{lllll}
    \includegraphics[width=2cm]{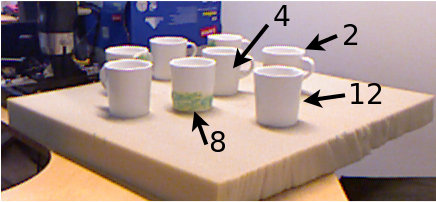} &
    \includegraphics[width=2cm]{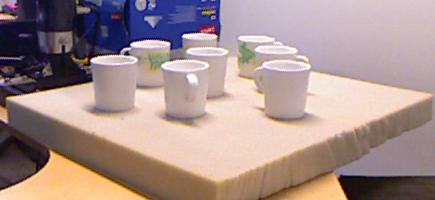} &
    \includegraphics[width=2cm]{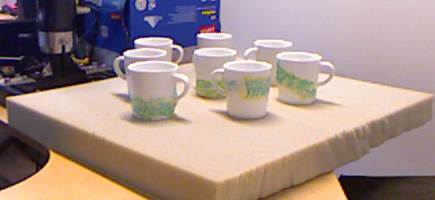} &
    \includegraphics[width=2cm]{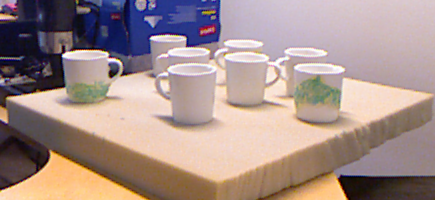} &
    \includegraphics[width=2cm]{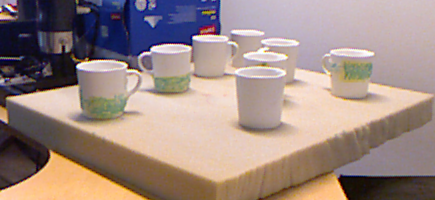} \\
    \includegraphics[width=2cm]{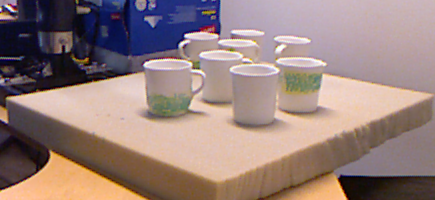} &
    \includegraphics[width=2cm]{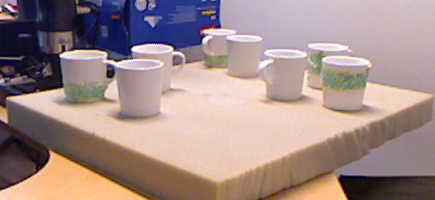} &
    \includegraphics[width=2cm]{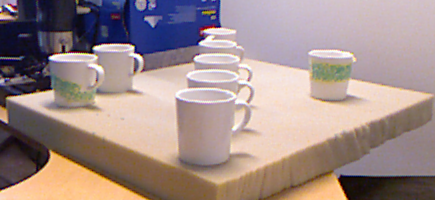} & 
    \includegraphics[width=2cm]{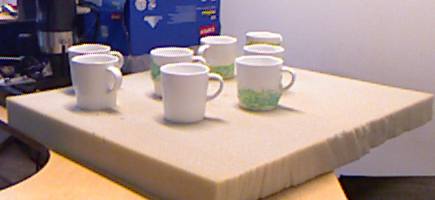} &
    \includegraphics[width=2cm]{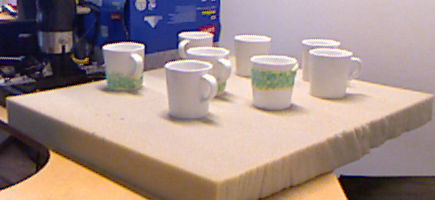}
  \end{tabular}
  \caption{Cropped kinect images of cup configurations used in the
    experiments. Each configuration contains four ``dirty'' (partly
    green color) and four ``clean'' cups. The green color on some cups
    is occluded.}
  \label{fig:simulation_point_clouds}
\end{figure}

In the simulated dynamics experiments, we used ten different captured
point clouds shown in Fig.~\ref{fig:simulation_point_clouds} as the
starting point for simulations. In these experiments, we form a world
model from the point cloud and then repeatedly sample an initial
belief and simulate the system using the probability model for 10 time
steps. To get an initial belief, we sample particles using the cup
dirtyness probability, which depends on past observations and which is
defined in Equation \ref{eq:attribute_probability} (dirtyness is an
object attribute). For evaluation purposes we also sample hidden
object specific grasp success probabilities. In more detail, we
sample for object $i$ the total amount of observed grasps $n_i =
n_i^{\textrm{succ}} + n_i^{\textrm{fail}}$ from a Gamma probability
distribution with shape $0.2$ and scale $5.0$, that is, a probability
distribution where small $n_i$ are common, but also large $n_i$ are
possible. We sample $n_i^{\textrm{succ}}$ from the uniform
distribution between $0$ and $n_i$, and keep $n_i^{\textrm{succ}}$ and
$n_i^{\textrm{fail}}$ constant during each simulation run. Note that
the magnitude of $n_i$ determines how much object specific grasp
properties affect the grasp success probability compared to occlusion.

\subsubsection{Results}

\begin{figure}[htb]
  \centering
  \includegraphics[width=8cm]{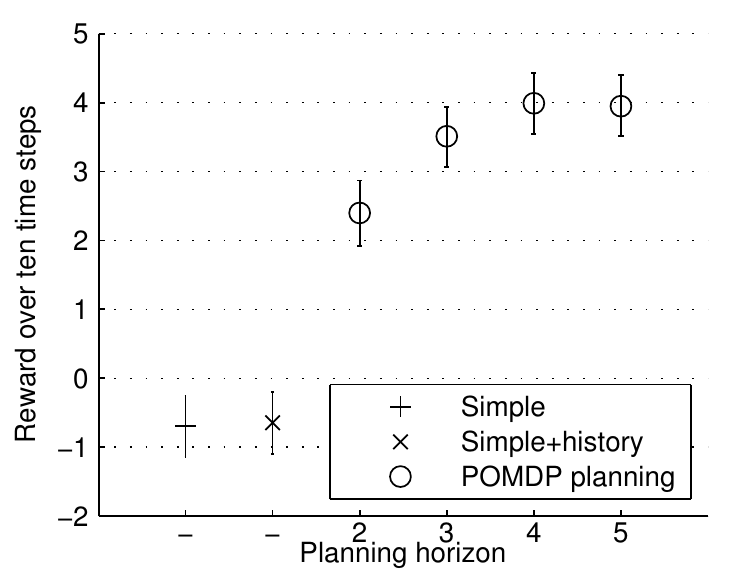}
  \caption{The average reward sum and its $95\%$ confidence interval
    (computed using bootstrapping) for the heuristic manipulation
    approach, heuristic manipulation approach utilizing grasp history
    information, and for the POMDP planning method.}
  \label{fig:simulation_results}
\end{figure}

Fig.~\ref{fig:simulation_results} compares POMDP planning with
different planning horizons, ranging from two to five, with the
heuristic manipulation approach. The POMDP policy graph had a width of
three. Fig.~\ref{fig:simulation_results} shows the average total
reward over $100$ simulation runs for each of the ten different cup
configurations shown in
Fig.~\ref{fig:simulation_point_clouds}. Overall, POMDP planning
achieves higher reward than the heuristic manipulation approach.
Interestingly, the performance difference between the heuristic
approach with and without grasp history is not significant. To study
this further, we ran over 2000 simulation runs for the scene shown in
the third image, upper row, in
Fig.~\ref{fig:simulation_point_clouds}. In this scene dirty cups are
in front and thus the heuristic approach can select between several
cups to move. Not surprisingly, the approach utilizing grasp history
performed better (with non-overlapping average reward confidence
intervals; not shown in Fig.~\ref{fig:simulation_point_clouds}).

\begin{figure}[htb]
  \centering
  \includegraphics[width=11cm]{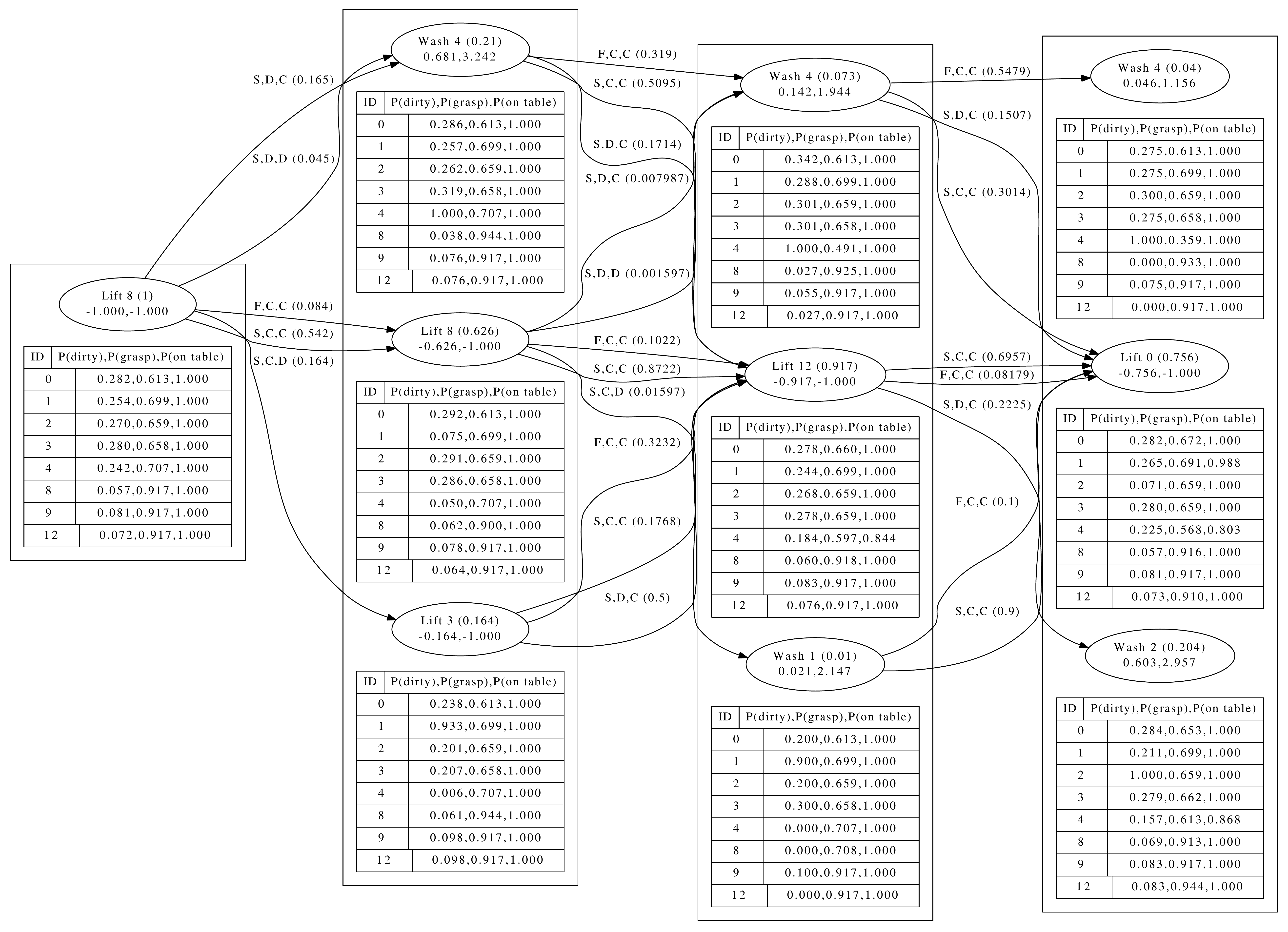}
  \caption{A policy graph optimized by the POMDP method for four time
    steps, when starting execution from the configuration shown in the
    first point cloud in Fig.~\ref{fig:simulation_point_clouds}. At
    each time step an agent executes the action associated with the
    current graph node, makes an observation, and moves to the next
    layer node along the corresponding observation edge. Each graph
    node shows its action, the visiting probability in parenthesis,
    the expected reward, and the expected reward divided by the
    visiting probability. Each graph edge is labeled with the
    observation, that is, three symbols, e.g.\ ``F,D,C'', and a
    visiting probability in parenthesis. The first observation symbol
    denotes grasp success (``S'') or failure (``F''); the second and
    third symbol denotes either dirty ``D'' or clean ``C'' for the
    first and second observed object, respectively. The box below a
    graph node displays for each object the dirtyness probability
    (``P(dirty)''), grasp success probability (``P(grasp)''), and the
    probability for the object to be on the table (``P(on
    table)''). Noteworthy: 1) failed grasps decrease the grasp
    probability, 2) lifting objects yields information about the
    dirtyness of objects behind them, 3) POMDP planning yields complex
    behavior.}
  \label{fig:experiments_policy_graph}
\end{figure}

It is also interesting that a POMDP planning horizon of three works
significantly better than a horizon of two. Intuitively, one could
imagine that short conditional plans, such as ``lift a cup, and then,
if the cup behind the lifted cup is dirty, move it into the
dishwasher'', would already perform very well. However, the results
suggest that many problems require a complex policy to gain high
reward. Fig.~\ref{fig:experiments_policy_graph} shows a compact policy
graph computed by the POMDP method for the first scene in
Fig.~\ref{fig:simulation_point_clouds}. The policy illustrates
information gathering through lifting cups, the effect of failed
grasps, and complex conditional planning. In the policy graph, the
agent lifts e.g.\ cups 8 and 12 (for reference, first RGB image in Fig.~\ref{fig:simulation_point_clouds} shows cups 2, 4, 8, and 12)
in order to gain information, and then
when observing cups 4 or 2 as dirty, tries to move them into the
dishwasher. In time step two, when the move of cup 4 into the
dishwasher fails, the grasp probability of cup 4 decreases. In time
step three, the agent tries to move cup 4 again. This highlights the
important feature of principled uncertainty handling in POMDP
planning. Even though grasping failed previously, the planner tries to
move the same cup, because compared to the alternatives the grasp
probability is still high enough.

\begin{figure}[htb]
  \centering
  \includegraphics[width=8cm]{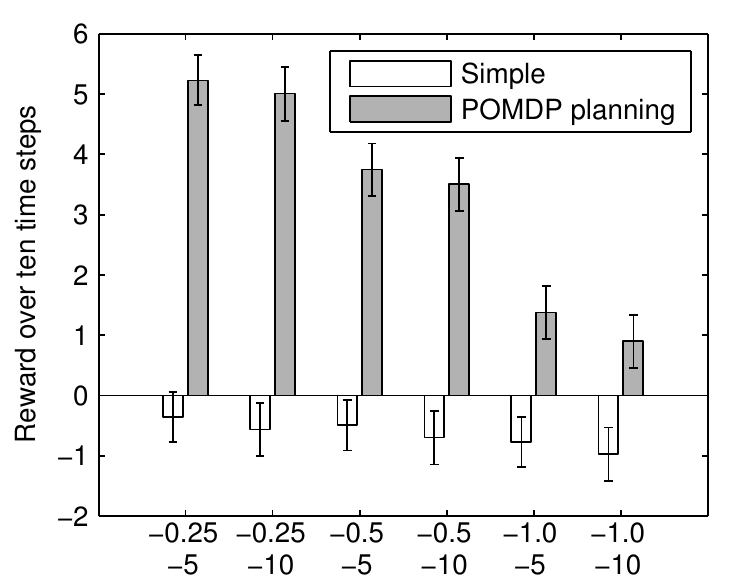}
  \caption{The average reward sum and its $95\%$ confidence interval
    (computed using bootstrapping) for the heuristic manipulation
    approach and for the POMDP planning method with a planning horizon
    of three for different reward scenarios. Each reward scenario has
    different rewards for moving a clean cup into the dishwasher ($-5$
    or $-10$), and for lifting a cup/a failed grasp attempt ($-0.25$,
    $-0.5$, or $-1.0$).}
  \label{fig:simulation_rewards}
\end{figure}

We also tested different reward
scenarios. Fig.~\ref{fig:simulation_rewards} shows performance for the
heuristic manipulation method and the POMDP method with a planning
horizon of three for different reward choices. In the experiment, we
varied the reward for lifting a cup/a failed grasp attempt and the
reward for putting a clean object into the dishwasher. The POMDP
method outperformed the heuristic method in each reward scenario. The
reward for failed grasps/lifting a cup had a significant effect on the
POMDP method's performance. One explanation is that when lifting cups
becomes more expensive the benefit of planning over complex
action-observation sequences decreases.

\subsection{Robot arm experiments}
\label{sec:robot_arm_experiments}

In the previous section, we simulated world dynamics using a world
model created from real robot grasps and point clouds captured by the
visual sensor. In this section, we present experiments with a physical
robot arm. In Section~\ref{sec:robot_arm_demonstrations}, we
demonstrate crucial parts of our world model. In
Section~\ref{sec:robot_arm_quantitative}, we compare quantatively the
performance of the greedy heuristic approach with the proposed POMDP
approach. In the demonstrations, we show the usefulness of information
gathering actions, such as lifting cups, in occluded
settings. Furthermore, we experimentally investigate when object
specific adaptive grasp probabilities are required. In addition, we
examine in which situations the heuristic manipulation approach
suffices for efficient operation, and when instead more comprehensive
POMDP based decision making is required. The quantative experiments
show that the POMDP based approach significantly outperforms the
simple greedy approach and yield insights, for instance, on why online
planning is beneficial. Overall, the experiments show that real world
problems require a model that takes occlusion into account, that
multi-object manipulation problems require multi-step POMDP planning,
and that adaptive action success probabilities are necessary in many
situations.

We performed robot arm experiments using the Kinova Jaco arm. In the
robot arm experiments, the Kinect sensor observes the scene, a method
decides which action to execute, and then the robot arm executes the
action. At each time step we estimate a belief from the captured point
cloud and add the observation history information to this belief, to
get the current belief. The method under evaluation decides on an
action using the current belief. In order to maintain a consistent
observation history and for detecting when a grasp succeeded or
failed, we match current objects to objects in the previous time step:
if an object is less than $4$cm from its last spatial position, we
assume it is the same object. If an object exists at the same location
after it was moved or lifted, we assume the grasp failed.

\subsubsection{Demonstrations}
\label{sec:robot_arm_demonstrations}

We claim that in multi-object manipulation, the robot may need to
perform information gathering actions when objects are occluded, or
when the grasp success probabilities of objects differ. However, when
objects are in plain sight and easy to grasp decision making is
easier. In this case, the problem requires no multi-step planning, and
the heuristic policy of moving all cups that appear dirty into the
dishwasher is sufficient. To test this, and to test whether our
observation and state space models are applicable in physical robot
arm experiments (we tested the model also in several other robot arm
experiments which are discussed below), we performed robotic
manipulation in a setup with dirty cups which are not
occluded. Fig.~\ref{fig:two_dirty_cups} shows how the heuristic
manipulation approach successfully moves the dirty cups into the
dishwasher in this setup.

To test our occlusion model, and to test whether occlusion requires
more complex decision making, we performed an experiment where the
dirtyness of a cup is not apparent because another cup partly
occludes the view on the dirty cup. The experiment in
Fig.~\ref{fig:dirty_cup_behind_clean_cup} demonstrates how the heuristic
manipulation approach does not consider information gathering, and
thus fails in the task. On the other hand, multi-step POMDP planning
takes into account that the dirty cup may in fact be dirty, even
though the robot observes it as clean, because the robot makes wrong
observations on occluded cups with a high probability. The POMDP
approach lifts the clean cup, gains new information on the dirty cup,
that is, observes the dirty cup as dirty, which increases the
probability of the cup being actually dirty, and then successfully
moves the dirty cup into the dishwasher.

\newlength{\myheight}
\settoheight{\myheight}{\hbox{\includegraphics[width=0.31\textwidth]{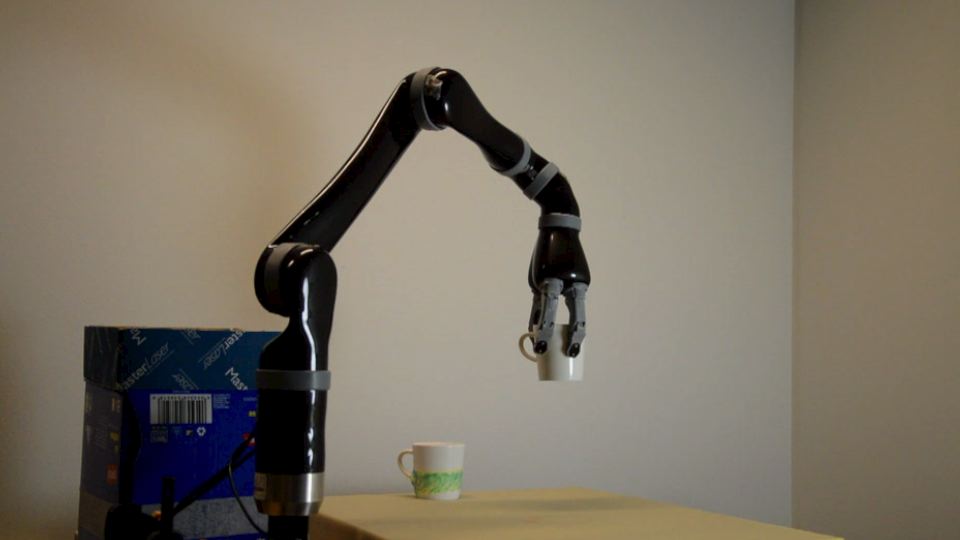}}}

\begin{figure}[htb]
  \centering
  \begin{subfigure}[b]{\textwidth}
    \centering
    \setlength{\tabcolsep}{1pt}
    \begin{tabular}{lllll}
      \includegraphics[height=\myheight]{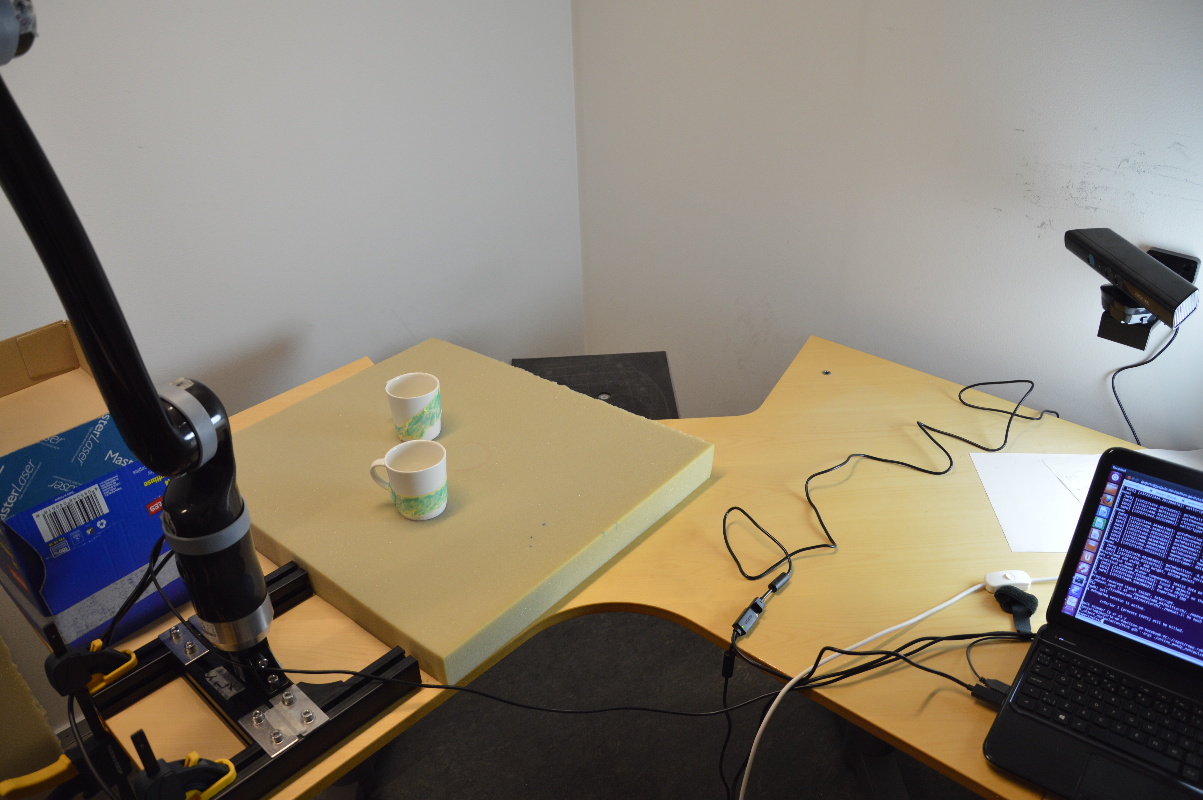} &
      \raisebox{0.5\myheight}{$\rightarrow$} &
      \includegraphics[width=0.31\textwidth]{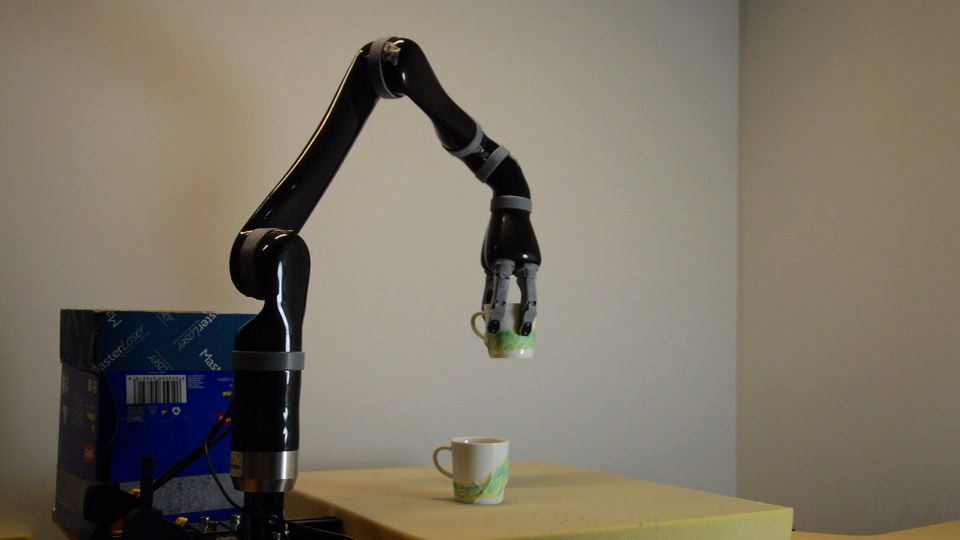} &
      \raisebox{0.5\myheight}{$\rightarrow$} &
      \includegraphics[width=0.31\textwidth]{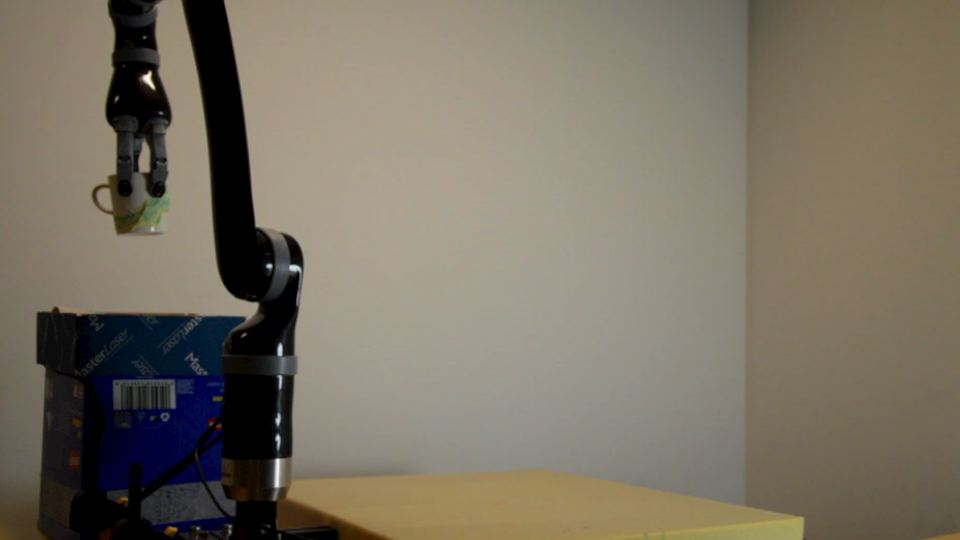}
    \end{tabular}
    \caption{The heuristic approach moves dirty cups which are not occluded into
      the dishwasher.}
    \label{fig:two_dirty_cups}
  \end{subfigure}\\
  \begin{subfigure}[b]{\textwidth}
    \centering
    \setlength{\tabcolsep}{1pt}
    \begin{tabular}{lllll}
      \smash{\raisebox{-0.5\height}{\includegraphics[height=\myheight]{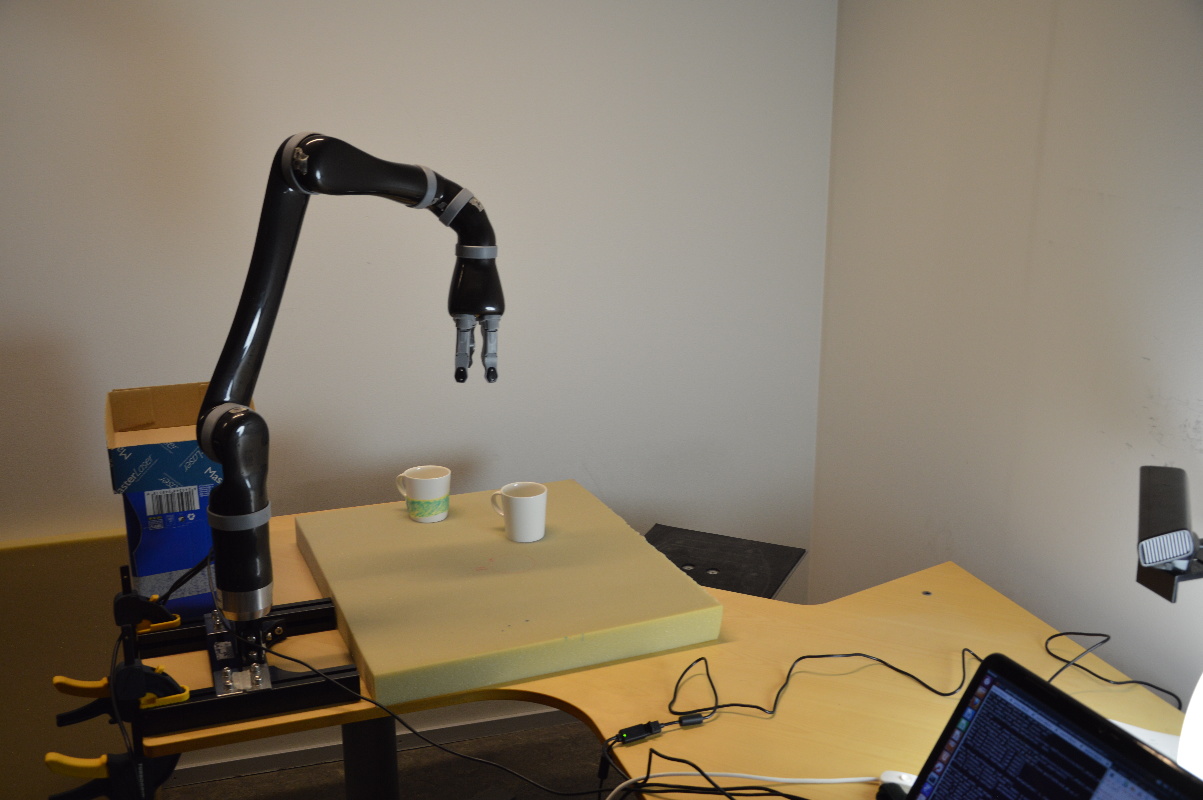}}} &
      \raisebox{0.4\myheight}{$\rightarrow$} &
      \includegraphics[width=0.31\textwidth]{dirty_cup_behind_clean_cup_pomdp1} &
      \raisebox{0.5\myheight}{$\rightarrow$} & 
      \includegraphics[width=0.31\textwidth]{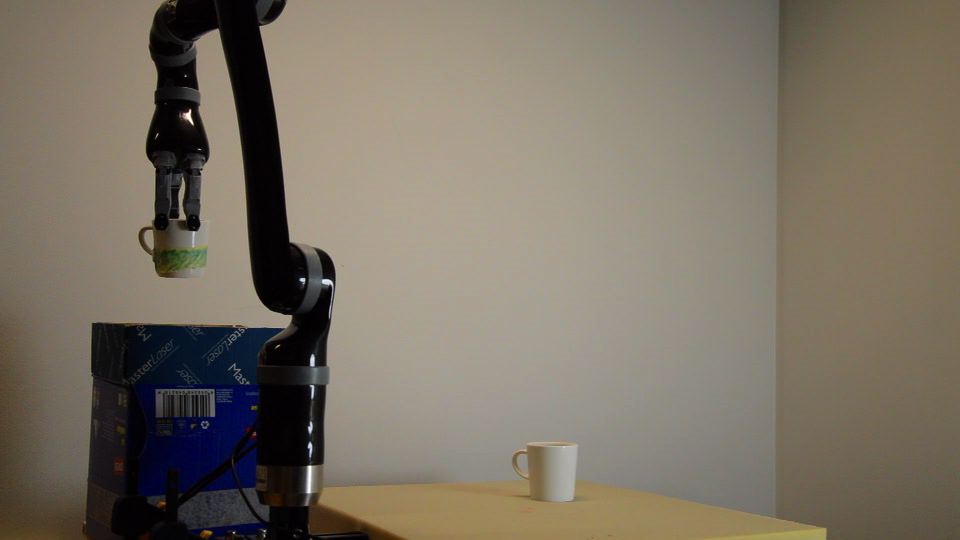} \\
      & \raisebox{0.6\myheight}{$\rightarrow$} &
      \includegraphics[width=0.31\textwidth]{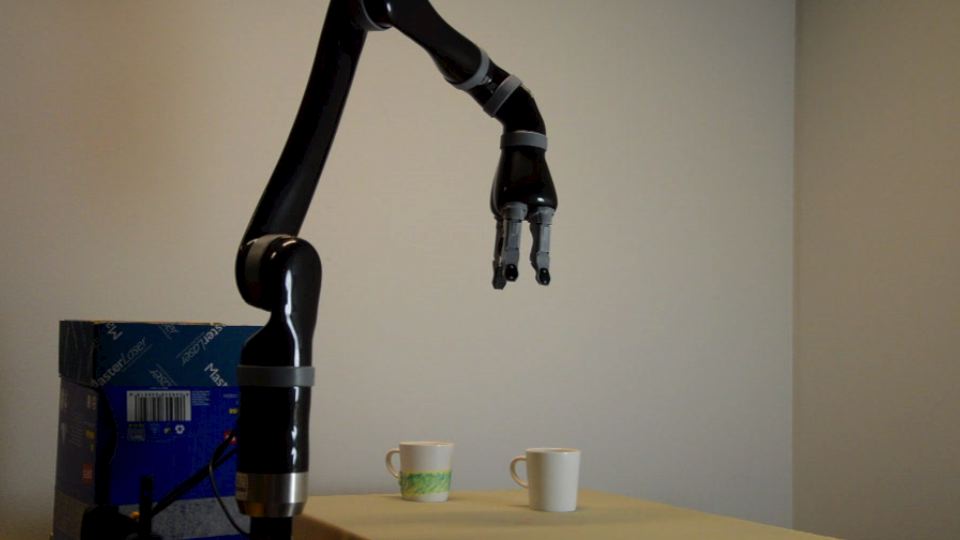} &
      \raisebox{0.5\myheight}{$\rightarrow$} & 
      \includegraphics[width=0.31\textwidth]{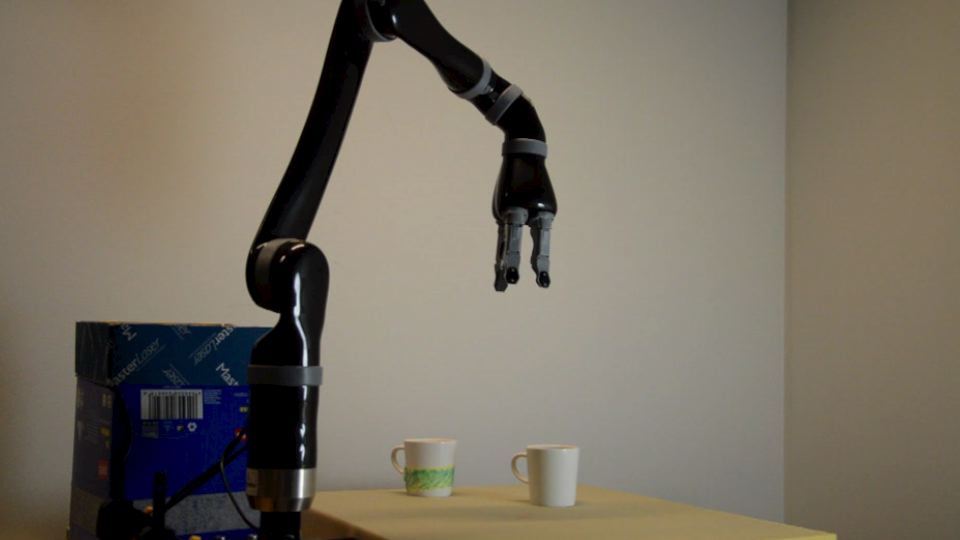}
    \end{tabular}
    \caption{Because of occlusion the robot observes a dirty cup as
      clean. \textbf{Top:} In order to gain more information, the
      POMDP approach lifts the occluding cup, and then, when observing
      the dirty cup correctly, moves it to the
      dishwasher. \textbf{Bottom:} The heuristic approach executes the
      Finish action, because all cups appear clean.}
    \label{fig:dirty_cup_behind_clean_cup}
  \end{subfigure}\\
  \caption{The robot tries to move possibly occluded dirty cups
    (partly green color) into the dishwasher (blue box).}
  \label{fig:occluded_cups}
\end{figure}

\begin{figure}[htb]
  \centering
  \begin{subfigure}[b]{\textwidth}
    \centering
    \setlength{\tabcolsep}{1pt}
    \begin{tabular}{lllll}
      \smash{\raisebox{-0.5\height}{\includegraphics[width=0.30\textwidth]{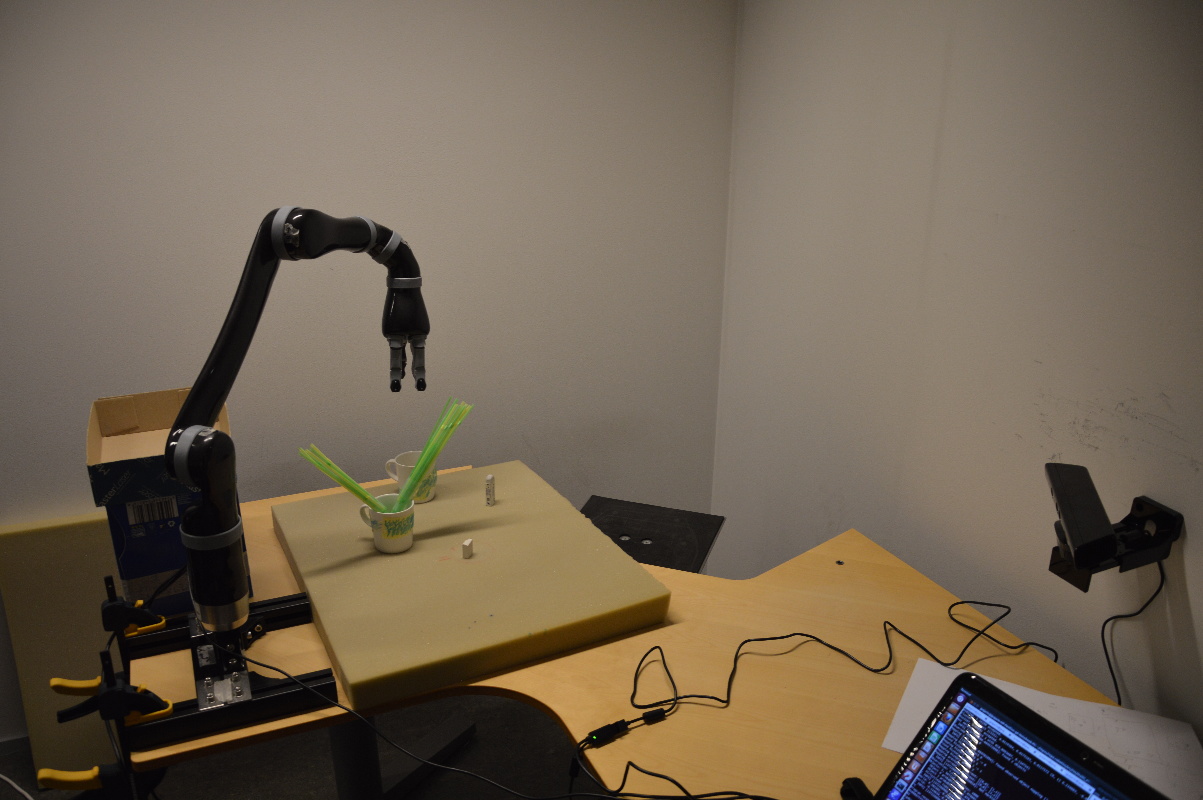}}} &
      \raisebox{0.4\myheight}{$\rightarrow$} &
      \includegraphics[width=0.30\textwidth]{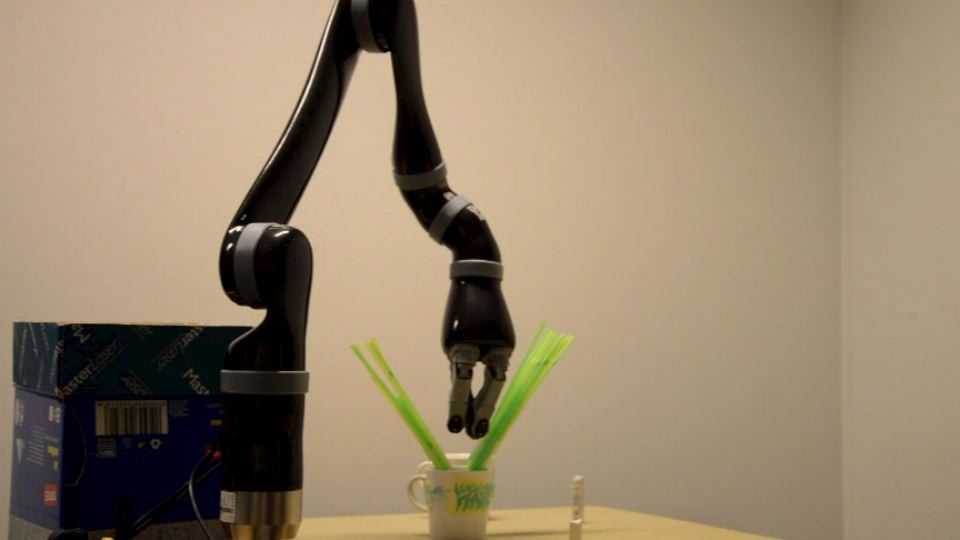} &
      \raisebox{0.5\myheight}{$\rightarrow$} & 
      \includegraphics[width=0.30\textwidth]{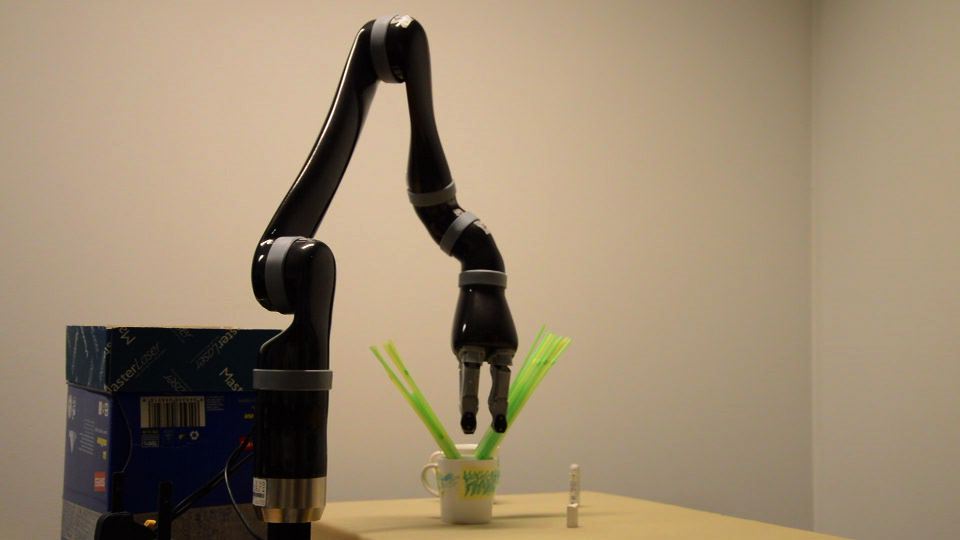} \\
      & \raisebox{0.6\myheight}{$\rightarrow$} &
      \includegraphics[width=0.30\textwidth]{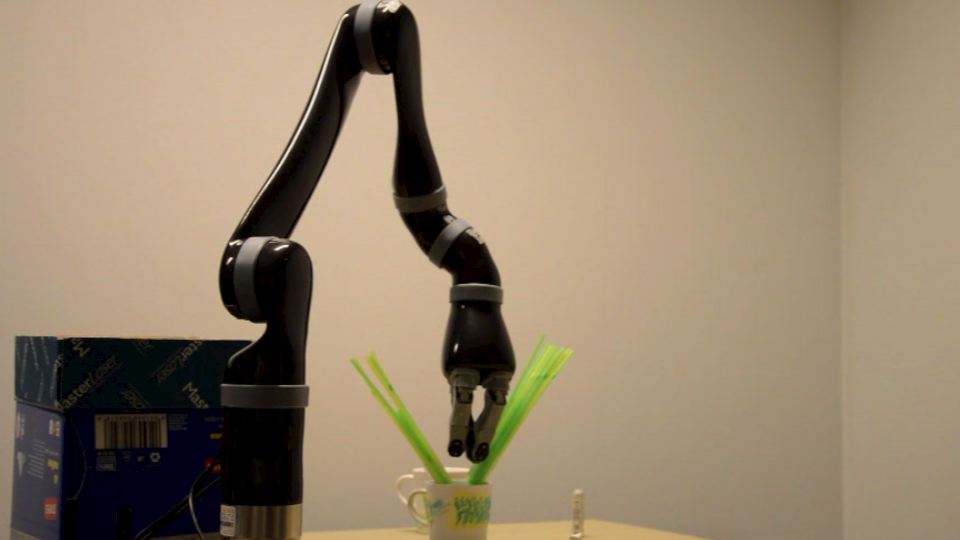} &
      \raisebox{0.5\myheight}{$\rightarrow$} & 
      \includegraphics[width=0.30\textwidth]{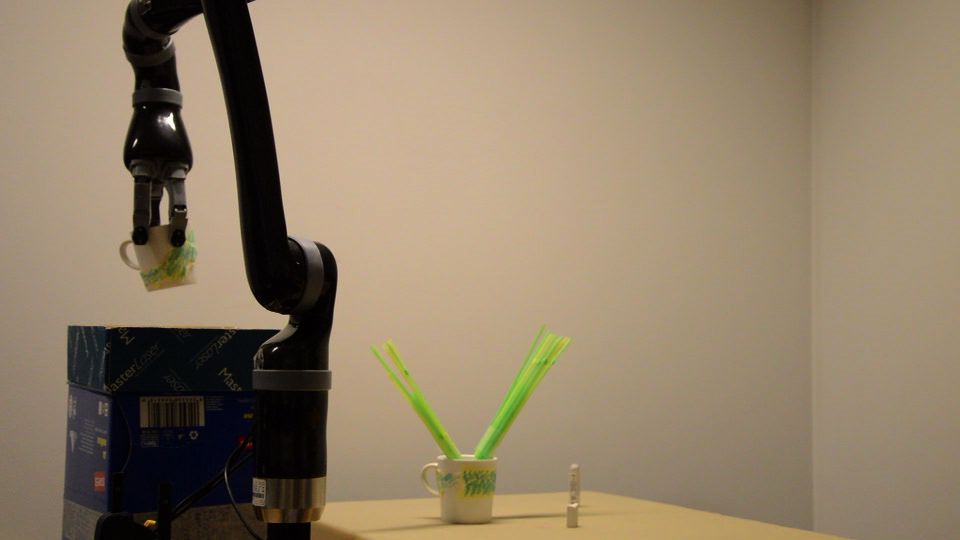}
    \end{tabular}
    \caption{The robot fails to grasp a dirty cup that contains
      drinking straws. \textbf{Top:} The heuristic approach which does
      not consider grasp history tries to move the same dirty cup
      again. \textbf{Bottom:} The heuristic approach which takes grasp
      history into account moves the other dirty cup into the
      dishwasher.}
    \label{fig:grasp_fails}
  \end{subfigure}\\
  \begin{subfigure}[b]{\textwidth}
    \centering
    \setlength{\tabcolsep}{1pt}
    \begin{tabular}{lllll}
      \includegraphics[width=0.30\textwidth]{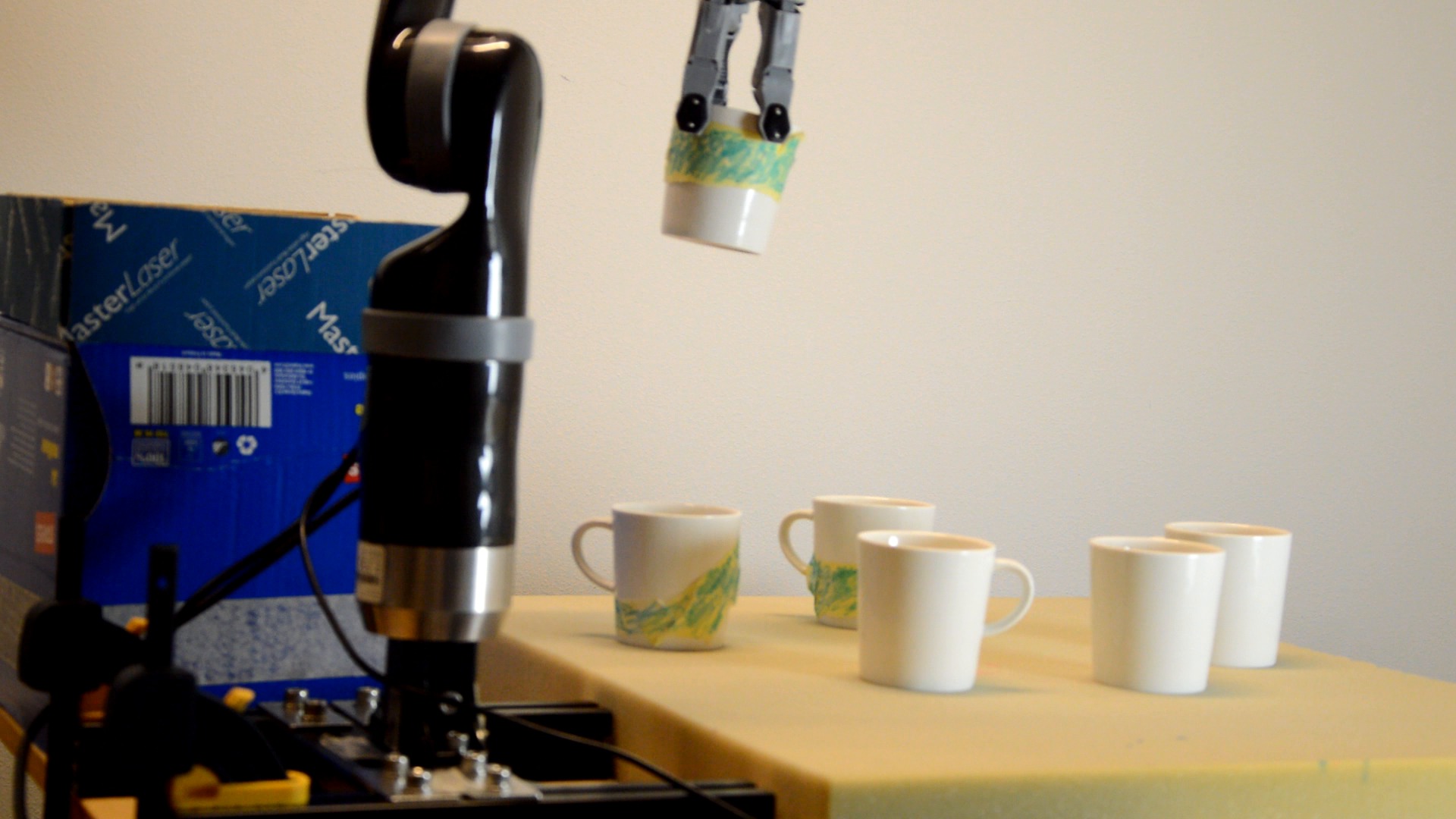} &
      \raisebox{0.5\myheight}{$\rightarrow$} &
      \includegraphics[width=0.30\textwidth]{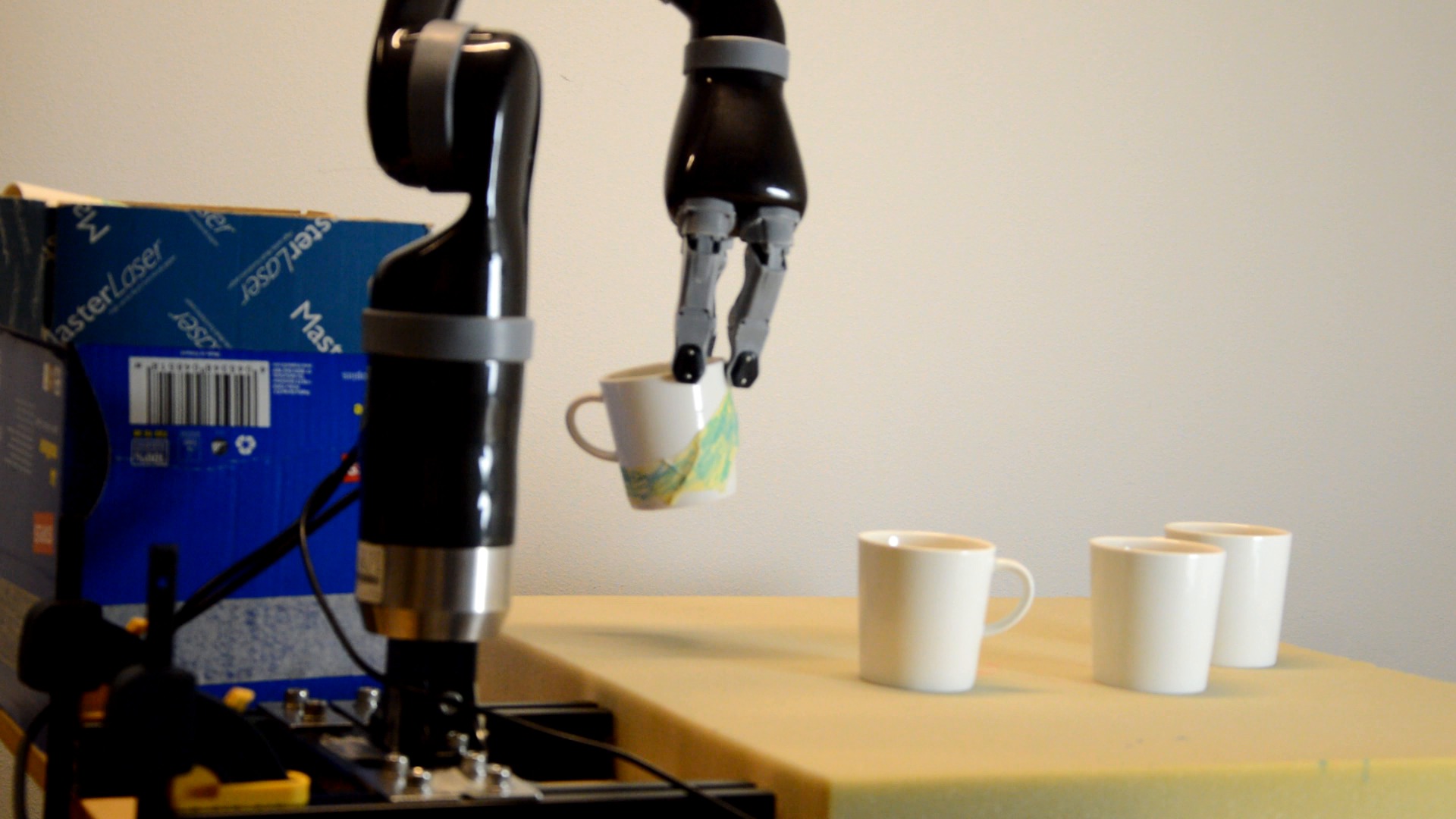} &
      \raisebox{0.5\myheight}{$\rightarrow$} &
      \includegraphics[width=0.30\textwidth]{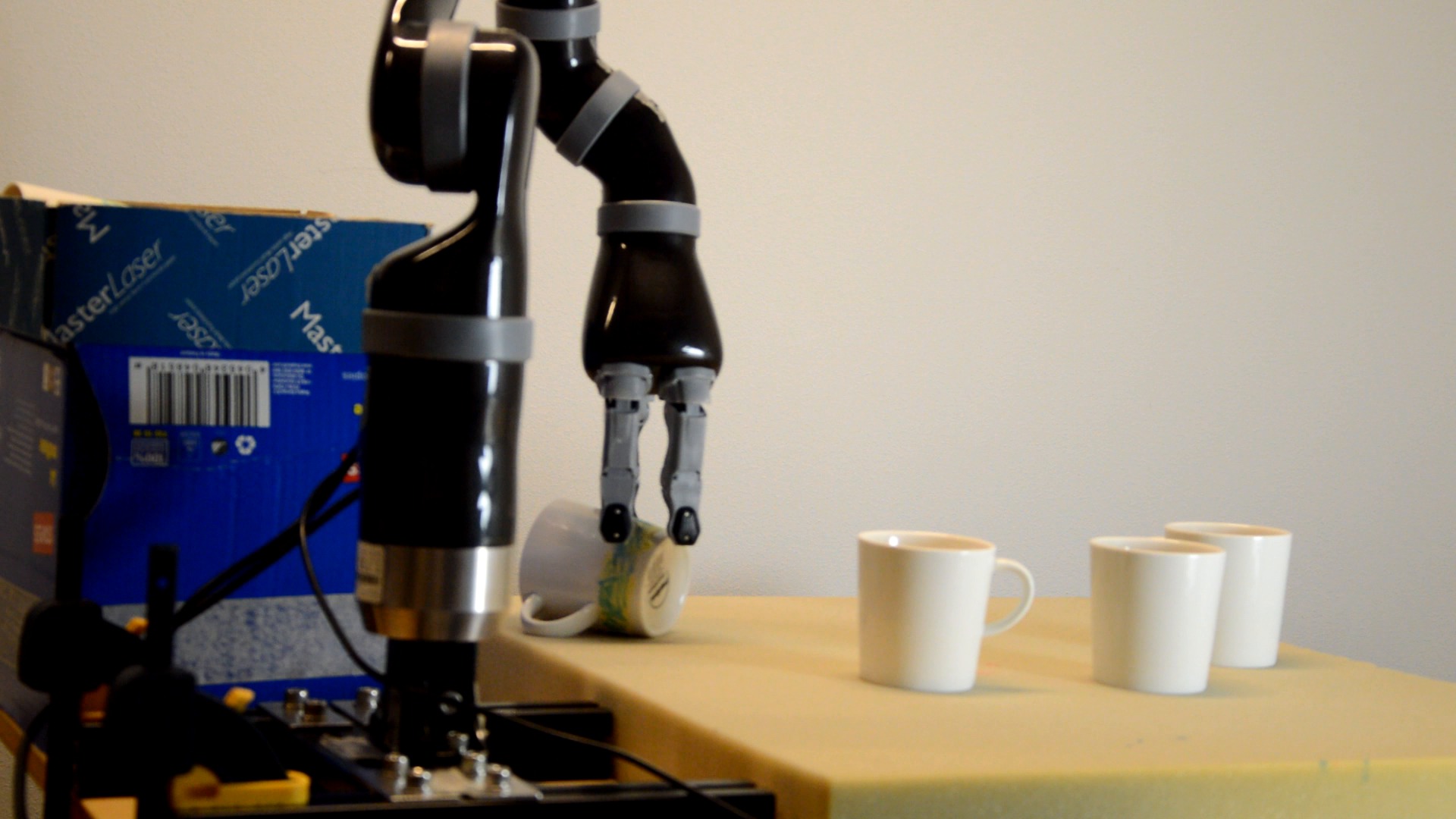}
    \end{tabular}
    \caption{Grasping becomes harder. The robot has moved two dirty
      cups into the dishwasher, when another dirty cup drops onto
      the table and remains resting on its side. Following grasp
      attempts fail, because the cup is now more difficult to grasp.}
    \label{fig:cup_falls_down}
  \end{subfigure}%
  \caption{The robot tries to move dirty cups (partly green color)
    into the dishwasher (blue box). Some of the cups are harder to
    grasp than others.}
  \label{fig:grasp_failure}
\end{figure}

Previously, we claimed that real world multi-object manipulation
problems require an object specific adaptive grasp success
probability. To test this claim and to verify that our adaptive grasp
success model works, we performed an experiment with two dirty cups
where the first cup is slightly occluded, and the second cup contains
drinking straws that make correct grasping more difficult. The robot
tries to move the second cup always first, because the occlusion on
the first cup makes the initial grasp success of the second cup
higher. For simplicity, we compared the heuristic manipulation approach
with and without adaptive grasp success probabilities. As shown in
Fig.~\ref{fig:grasp_fails}, both methods fail to grasp the second cup
because of the drinking straws. The adaptive grasp success probability
method updates the grasp success probability after observing a failed
grasp, and moves the first dirty cup successfully into the
dishwasher. The method that does not take grasp success history into
account tries to grasp the same second cup again, even though an
easier to grasp dirty cup would be available. These kind of situations
occur often in practice. During experimentation with the robotic arm
for example, as shown in Fig.~\ref{fig:cup_falls_down}, the robot
moves dirty cups, but when it moves the third dirty cup, the cup falls
down and remains in a harder to grasp pose. We observed that when
further grasps on the object failed, the grasp success probability
decreased as expected.

\subsubsection{Quantitative results}
\label{sec:robot_arm_quantitative}

\begin{figure}[htb]
  \setlength{\tabcolsep}{5pt}
  \begin{tabular}{llll}
    \includegraphics[width=2.6cm]{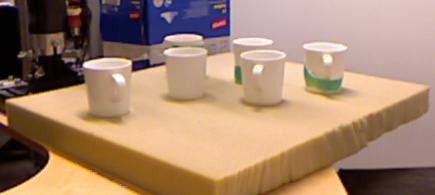} &
    \includegraphics[width=2.6cm]{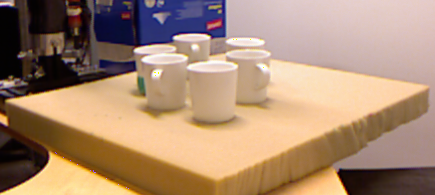} &
    \includegraphics[width=2.6cm]{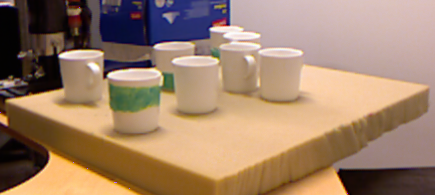} &
    \includegraphics[width=2.6cm]{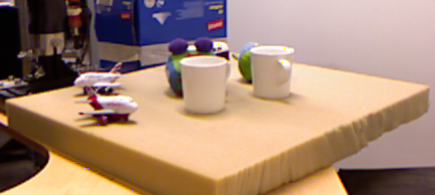} \\
    \includegraphics[width=2.6cm]{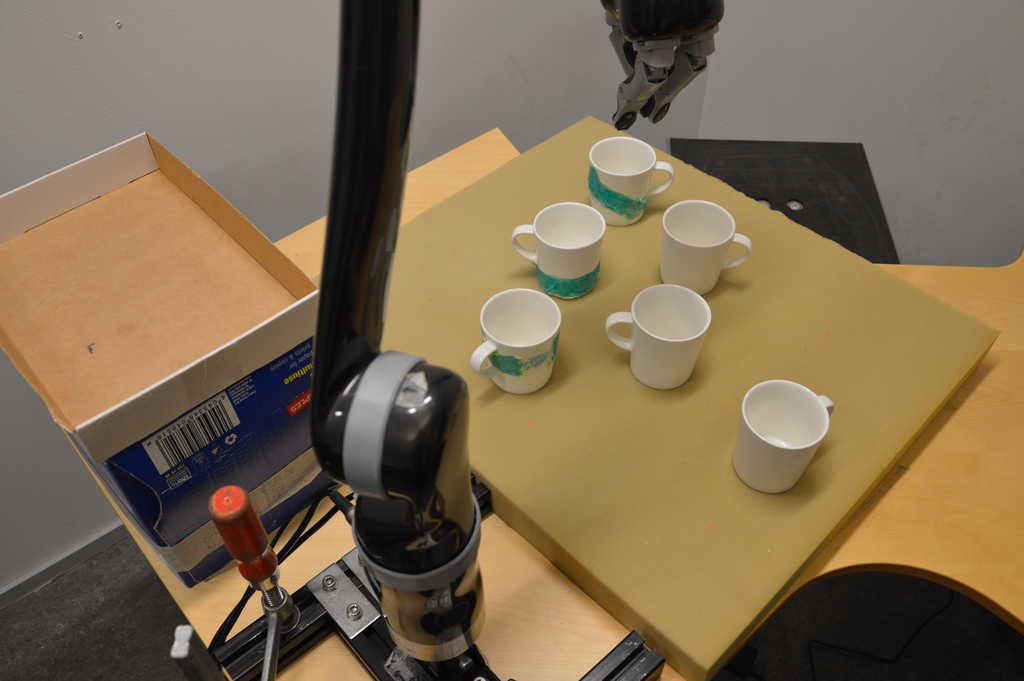} &
    \includegraphics[width=2.6cm]{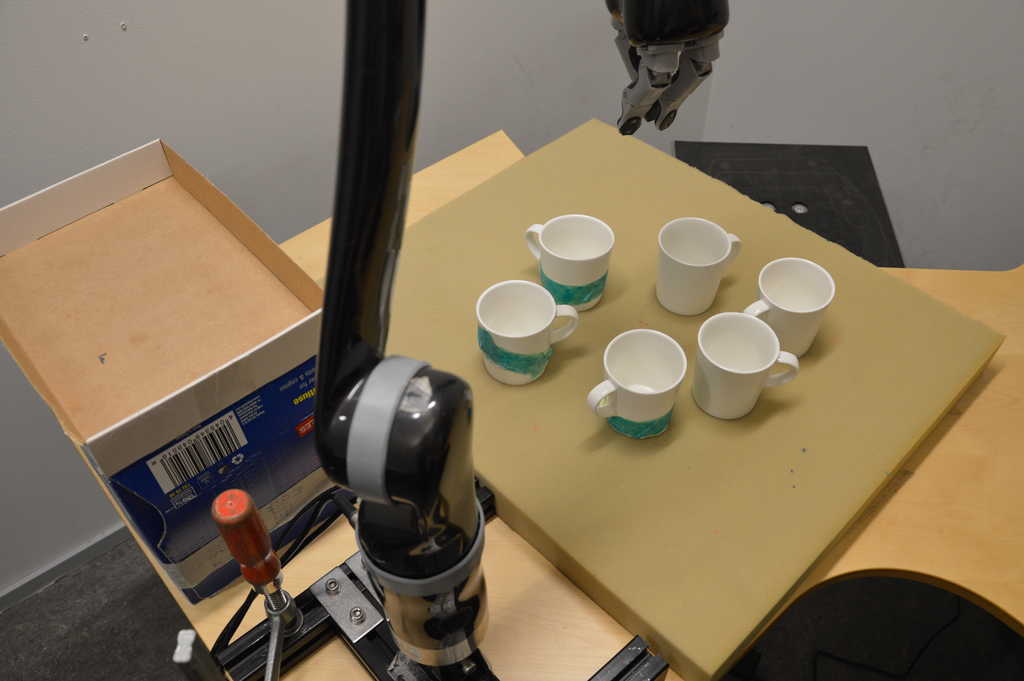} &
    \includegraphics[width=2.6cm]{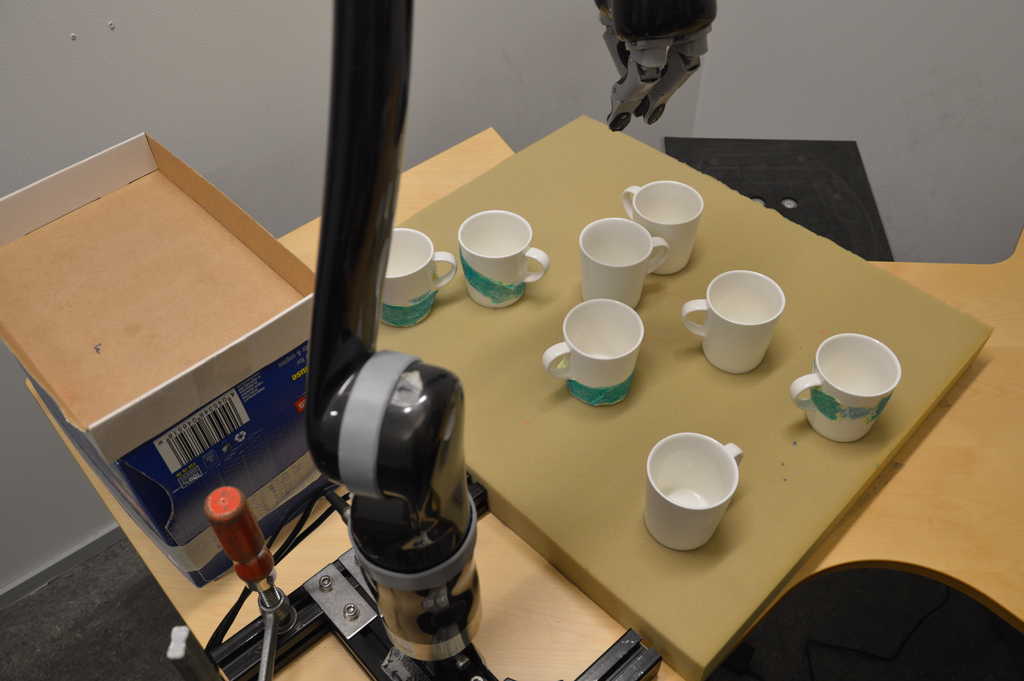} &
    \includegraphics[width=2.6cm]{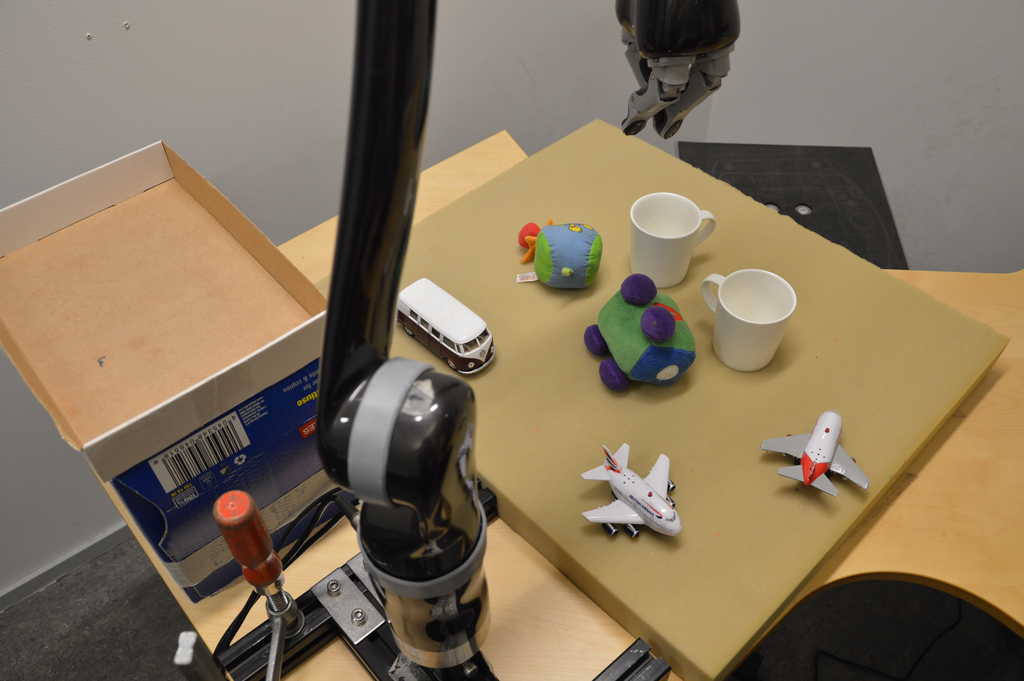} \\
  \end{tabular}
  \caption{\textbf{Top row}: cropped kinect images of the four scenes
    used in the robot arm experiments. \textbf{Bottom row}:
    corresponding photographs of the scenes. Each scene contains
    ``dirty'' (partly green color) and ``clean'' objects. Scenes one
    to three contain only cups but scene four contains also several
    toys.}
  \label{fig:physical_point_clouds}
\end{figure}

\begin{figure}[htb]
  \centering
  \begin{minipage}[t]{0.4\linewidth}
    \vspace{0pt}
    \hspace{-1em}
    \includegraphics[width=6cm]{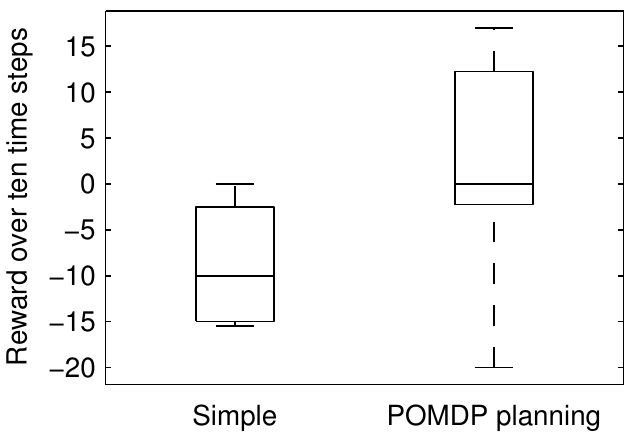}
  \end{minipage}%
  \begin{minipage}[t]{0.5\linewidth}
    \vspace{1em}
    \centering%
    \hspace{1em}
    \footnotesize
    \begin{tabular}{|c|r|r|}%
      \hline
      Scene & Simple & POMDP \\\hline
      \#1 & $-9.1$ & $11.6$ \\\hline
      \#2 & $-15$ & $-4$ \\\hline
      \#3 & $0$ & $2.6$ \\\hline
      \#4 & $-10$ & $-1.7$ \\\hline
    \end{tabular}
  \end{minipage}
  \caption{Robot arm experiments. We executed both the simple greedy
    approach ``Simple'' and the POMDP approach with a planning horizon
    of three ``POMDP planning'' five times in each of the four scenes
    shown in Fig.~\ref{fig:physical_point_clouds}. \textbf{Left:}
    boxplot of the reward over ten time steps. ``POMDP planning'' was
    significantly better than ``Simple'' (the $p$-value was $0.00059$
    in the Mann-Whitney $U$ test \cite{mann47}). \textbf{Right:} for
    both methods the average reward in each scene. ``POMDP planning''
    had a larger average reward in each scene.}
  \label{fig:physical_results}
\end{figure}

In addition to the demonstrations, we performed a quantitative
comparison between the simple greedy heuristic approach and the
proposed POMDP approach in physical robot arm experiments. Similar to
the experiments with simulated dynamics in
Section~\ref{sec:simulated_dynamics}, the goal was to move dirty, that
is, partly green objects, into a ``dishwasher''. An object was
observed dirty if the number of green pixels was at least $100$.

Fig.~\ref{fig:physical_point_clouds} shows the four different scenes
used. The fourth scene contains also toys to demonstrate the
genericity of our approach. In each scene, we placed the objects on
the table, and then ran the simple heuristic method and the POMDP
method with a planning horizon of 3 after each other, five times each,
yielding a total of twenty runs for each method over all four
scenes. We reconstructed a scene after each
run. Fig.~\ref{fig:physical_results} shows the results. Overall, the
POMDP approach significantly outperformed the heuristic
approach. Moreover, in each scene, the POMDP approach received higher
rewards on average. Regarding planning times, on a single low
performance AMD A10-4600M CPU core the heuristic approach took roughly
$0.02\%$ ($2$ milliseconds per time step) and the POMDP approach took
$4.7\%$ ($2.6$ seconds per time step) of the total execution time. The
planning time for both methods was negligible compared to the time
sensor processing and moving the robot arm required.

Performance wise the heuristic approach was closest to the POMDP
approach in scene 3. In scene 3, the two partly green objects closest
to the Kinect were easy to grasp and the heuristic approach always
successfully moved them to the dishwasher. Because of heavy occlusion
the two partly green objects farthest from the Kinect were very hard
to grasp. Therefore, while being usually able to move the easy to
grasp objects, the POMDP approach had more difficulty in moving the
other two partly green objects. Interestingly, among individual
experiment runs, the POMDP approach had both the lowest ($-20$) and
highest ($17$) reward. The lowest reward was possible because of the
grasp and observation uncertainty, and because the POMDP approach was
more active than the heuristic approach. Another interesting
observation from the experiments was that occasionally an object could
be dropped or tipped over. Our POMDP model does not explicitly take
these kinds of events into account. However, in spite of this, the
POMDP approach adapted to these unexpected situations because it
always planned actions based on the belief estimated from current
sensor readings.

\subsection{Discussion}

The experiments confirm that multi-step POMDP planning is useful, when
the order of actions is critical to the successful completion of the
task. In particular, a POMDP estimates the value of information
optimally. In an uncertain world, the probabilistic model used in
POMDPs can weight different action choices in a principled manner. In
contrast to a greedy approach, a POMDP may select actions that gather
information, but do not yield immediate reward, when the problem so
requires. In the multi-object manipulation experiments, the robot had
to decide between lifting objects to gather information or moving
objects that appear dirty into the dishwasher. Our POMDP model
includes grasping success and learns grasping probabilities. Grasping
unknown objects requires object specific grasp probabilities because
each object may be different. However, even when predefined object
models are available, adaptive object specific grasp probabilities may
be useful; especially in heavily cluttered settings, with multiple
objects, the large uncertainty about object pose and identity make
grasping some objects harder than others and requires an adaptive
approach.

\section{Conclusion}

We presented a POMDP model for multi-object manipulation of unknown
objects in a crowded environment. Because objects are occluded, their
attributes are harder to observe and they are harder to manipulate. To
address this, our POMDP model uses an \emph{occlusion ratio}
to define how much an object occludes another one. We use the
occlusion ratio as a parameter in the observation and grasp
probabilities of objects. In addition to occlusion specific grasp
probabilities, our model also includes automatically adapting object
specific grasp probabilities. To compute compact policies for the
computationally complex POMDP model, we presented a new POMDP method
that optimizes a policy graph using particle filtering. The method
allows multi-step POMDP planning, both offline and online.

Experiments confirm that a heuristic greedy manipulation approach is
not adequate for multi-object manipulation, but instead, the problem
requires complex conditional multi-step POMDP plans that take long
term effects into account. Moreover, object specific grasp
probabilities are needed in many real-world situations.

In the future we plan to apply the presented POMDP model to other
kinds of robotic tasks. Currently, we are extending the POMDP model to
take into account the uncertainty in the composition of objects from
segments. In general, to obtain true long-term autonomy, we believe
that a robot should base its decisions on prior learned knowledge and
adjust its world model to the specific environment it operates in. For
this purpose a probabilistic Bayesian framework should be used that
allows the robot to operate and learn in an uncertain, unstructured
environment. In contrast to engineered solutions, learning offers the
possibility to find solutions that generalize to unexpected situations
and a possibility for autonomous adaptation. Our goal is an autonomous
robot which can be placed in a complex new environment and which then
knows how to adapt to the new environment. The work presented here is
a step towards that goal.

\section*{Acknowledgements}

This work was supported by the Academy of Finland, decision 271394.

\bibliographystyle{plain}
\bibliography{long_robotic_planning}

\end{document}